\documentclass[10pt,twocolumn,letterpaper]{article}

\usepackage{cvpr}              

\usepackage[accsupp]{axessibility}

\usepackage{algorithm}
\usepackage{algorithmic}
\usepackage{newfloat}
\usepackage{listings}
\usepackage{amsmath}
\usepackage{multirow}
\usepackage{booktabs}
\usepackage{colortbl}
\usepackage{adjustbox}
\usepackage{xspace}
\usepackage{tabularx}
\usepackage{array}
\usepackage{minitoc}
\usepackage{titletoc}

\newcommand{\methodfullname}{\textbf{\textit{Adaptive Frame-Pruning (AFP)}}\xspace}
\newcommand{\methodshort}{\textbf{\textit{AFP}}\xspace}
\newcommand{\methodgraph}{\textbf{\textit{AFP+Graph}}\xspace}
\newcommand{\methodgraphspace}{\textbf{\textit{AFP + Graph}}\xspace}

\newcommand{\basevs}{\textit{VSLS}\xspace}
\newcommand{\baseaks}{\textit{AKS}\xspace}
\newcommand{\basetstar}{\textit{T*}\xspace}

\newcommand{\baseaksstar}{\textit{AKS*}\xspace} 
\newcommand{\basevsstar}{\textit{VSLS*}\xspace}

\newcommand{\lvb}{\textsc{Long\allowbreak VideoBench}\xspace}
\newcommand{\mme}{\textsc{VideoMME}\xspace}
\newcolumntype{C}{>{\centering\arraybackslash}X}

\usepackage{microtype}

\usepackage{enumitem}
\setlist{nosep} 

\definecolor{cvprblue}{rgb}{0.21,0.49,0.74}
\usepackage[pagebackref,breaklinks,colorlinks,allcolors=cvprblue]{hyperref}

\def\paperID{46397} 
\def\confName{CVPR}
\def\confYear{2026}

\title{Less is More: Token-Efficient Video-QA via Adaptive Frame-Pruning and Semantic Graph Integration}

\author{
Shaoguang Wang$^{1}$ \quad
Weiyu Guo$^{1}$ \quad
Ziyang Chen$^{1}$ \quad
Yijie Xu$^{1}$ \quad
Xuming Hu$^{1,2*}$ \quad
Hui Xiong$^{1,2}$\thanks{Corresponding authors.} \\[2mm]
$^1$The Hong Kong University of Science and Technology (Guangzhou), China\\
$^2$The Hong Kong University of Science and Technology, Hong Kong SAR, China\\[1mm]
{\tt\small swang440@connect.hkust-gz.edu.cn, xuminghu@hkust-gz.edu.cn, xionghui@ust.hk}
}

\begin{document}
\maketitle
\begin{abstract}
The practical application of Multimodal Large Language Models (MLLMs) to Video Question Answering (Video-QA) is severely hindered by the high token cost of processing numerous video frames. While keyframe selection is the dominant strategy for mitigating this, we identify a critical flaw: even state-of-the-art selectors produce prompts suffering from significant temporal redundancy, a challenge unique to video that we term `visual echoes'. This issue leads to context dilution and can paradoxically degrade performance. To address this dual challenge, we propose a novel refinement framework that synergistically combines \methodfullname with a lightweight text-based semantic graph. \methodshort intelligently prunes `visual echoes' by adaptively clustering frames, while the semantic graph provides crucial, low-cost semantic compensation. Conducting extensive experiments on the \lvb and \mme benchmarks against multiple state-of-the-art selectors, our approach demonstrates a drastic reduction in total input tokens by up to 82.2\%. Crucially, by creating a concise, high-quality prompt, our framework not only enhances efficiency but also demonstrates a remarkable ability to robustify and improve the accuracy of upstream selectors, achieving results that are highly competitive with, and often superior to, baselines that use vastly more frames. Code is publicly available at \url{https://github.com/shaoguangwang/Adaptive-Frame-Pruning}.
\end{abstract}    
\section{Introduction}

\begin{figure}[t]
\centering
\includegraphics[width=\columnwidth]{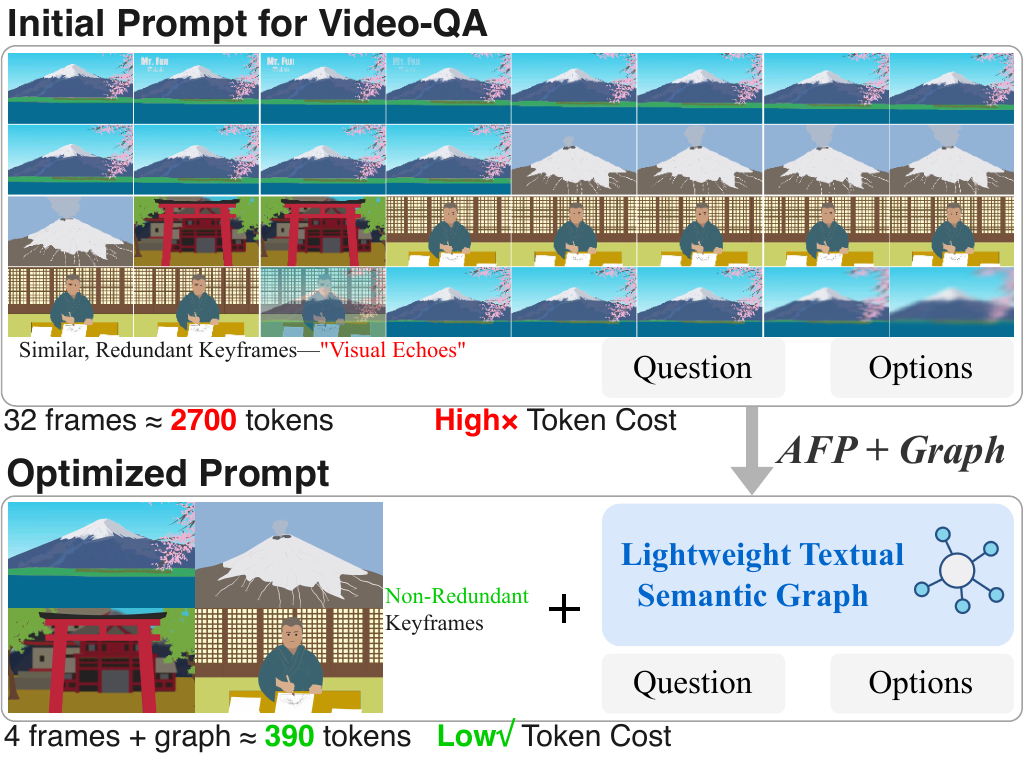} 
\caption{\textbf{Conceptual Overview of our Refinement Framework.} (a) An initial prompt from an upstream selector contains numerous `visual echoes' and has a high token cost. (b) Our framework refines this by pruning redundant frames with \methodshort and compensating with a semantic graph, resulting in an optimized, low-cost prompt that leads to the correct answer. \textit{Token counts shown are approximate illustrations based on OpenAI's pricing model; exact values vary by dataset and are detailed in Section~\ref{sec:exp_setup}.}}
\label{fig:concept}
\end{figure}

\begin{figure}[t]
\centering
\includegraphics[width=\columnwidth]{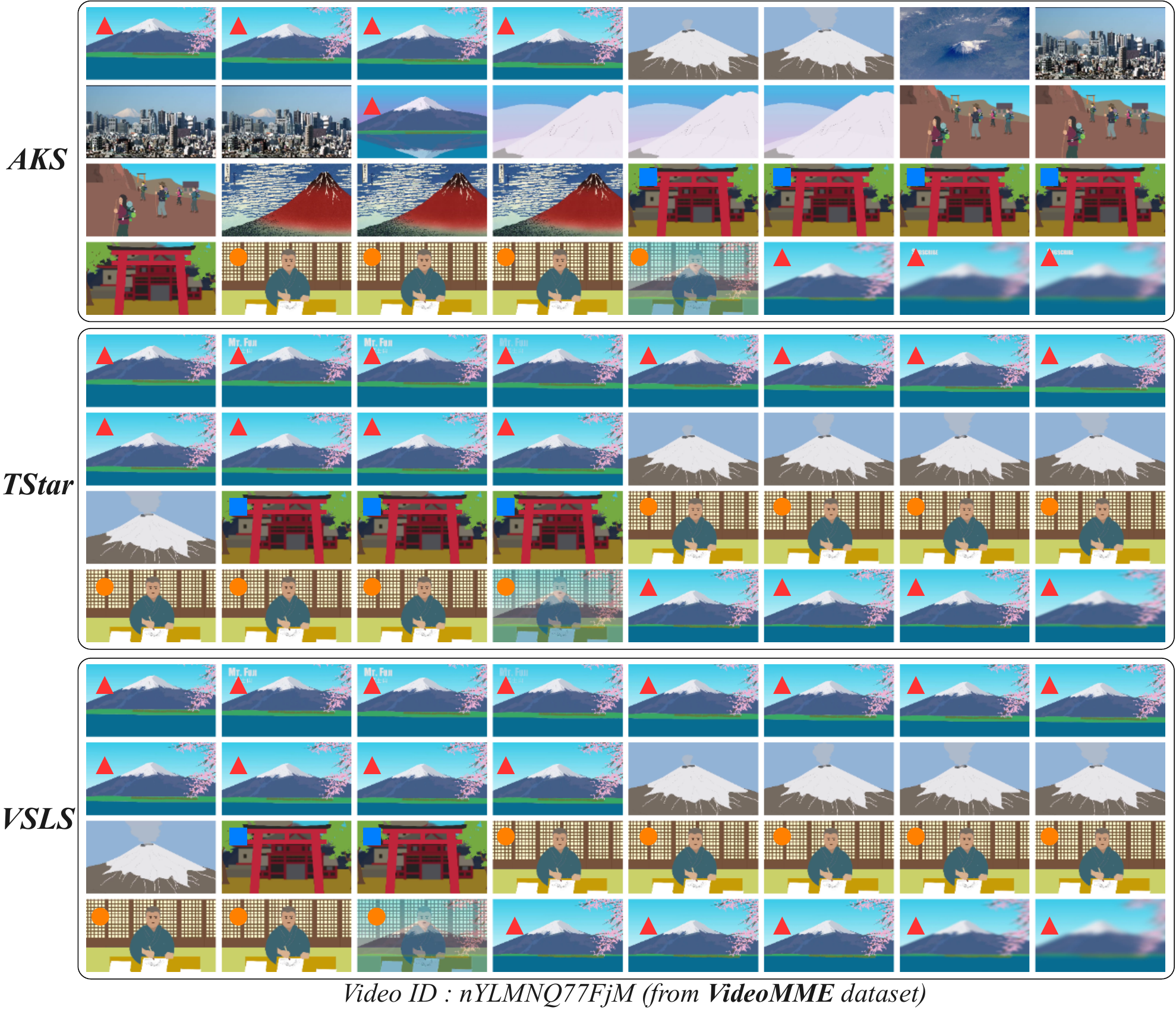} 
\caption{\textbf{``Visual Echoes'' are a Prevalent Issue Across Mainstream Keyframe Selectors.} We visualize 32 keyframes selected by three SOTA methods for a Video-QA task involving narrative understanding. All selectors exhibit severe redundancy, producing multiple near-identical frames for iconic subjects. We use colored bounding boxes to highlight these clusters of ``visual echoes,'' such as the Fuji mountain view, the seated man, and the red Torii gate.}
\label{fig:visual_echoes_comparison}
\end{figure}

The advent of Multimodal Large Language Models (MLLMs), such as GPT-4o, Gemini, Video-ChatGPT, MovieChat and Video-LLaVA~\cite{hurst2024gpt, team2023gemini, maaz2024video, song2024moviechat, lin2024video}, has led to a significant advancement in vision–language understanding. These models show strong ability in handling complex reasoning over visual and textual inputs, with Video Question Answering (Video-QA) becoming an important application. However, their use in the video domain is limited by the very high token cost. This issue is critical for long videos; for example, sampling a one-hour video at 1~fps produces 3{,}600 frames, resulting in a token count that far exceeds the context window of current MLLMs. This makes naive frame sampling ineffective and highlights the need for more selective frame sampling strategies.

This computational barrier forces a departure from naive sampling strategies. While dense uniform sampling is infeasible, simply resorting to a sparse, fixed sampling (e.g., 32 frames for a long video) is equally problematic. Such a \emph{query-agnostic} approach is fundamentally unreliable, as it runs a high risk of missing the brief but decisive moments essential for answering the specific question posed.

Given these constraints, \emph{query-aware} keyframe selection has emerged as the dominant preprocessing paradigm. Yet, this approach introduces a critical trade-off. To ensure no relevant information is missed (i.e., to maximize recall), state-of-the-art selectors are designed to be comprehensive, often returning a generous and often redundant set of candidate frames for downstream analysis. 

This inherent design choice leads to two significant, often-overlooked challenges. First, due to the temporal coherency of video, the selected keyframes inevitably contain significant redundancy, which we term \textbf{`visual echoes'}: temporally proximate frames with high visual similarity (illustrated conceptually in Figure~\ref{fig:concept}). For additional examples demonstrating this issue across diverse video types, including scientific animations and real-world footage, please see supplementary material. This tendency is not an isolated flaw of a particular method, but a widespread issue across the current paradigm, as we demonstrate in Figure~\ref{fig:visual_echoes_comparison} where multiple leading selectors exhibit this behavior. Second, the overabundance of visual information, even if relevant, leads to \textbf{`context dilution'}, which can introduce noise and overwhelm the MLLM's reasoning capacity. These twin challenges hinder the potential for a ``less is more'' paradigm, often leading to a counter-intuitive outcome where more frames do not necessarily yield better results, as noted in recent literature~\cite{ma2025drvideo}.


These observations lead us to a more refined research question: \textit{How can we \textbf{refine} a redundant and noisy set of initial keyframes into an optimal, token-efficient prompt that is both visually concise and semantically complete?}

To answer this question, we propose a novel, two-pronged framework that synergistically combines \textbf{\methodfullname} and a \textbf{text-based semantic graph}. As illustrated in our conceptual overview (Figure~\ref{fig:concept}), our framework operates as a universal refinement layer. The \methodshort module directly tackles the `visual echoes' problem by employing an adaptive clustering algorithm to merge redundant frames into a concise visual foundation. Concurrently, the lightweight semantic graph compensates for potential information loss that occurs during pruning. It provides a structured, low-cost summary of key entities and their relationships, ensuring that essential semantic context is preserved and highlighted for MLLMs. Together, they transform a high-cost, cluttered prompt into a highly efficient and effective one. To summarize, the main contributions of our work are threefold:
\begin{itemize}
    \item We formally define and demonstrate the prevalence of \textbf{`visual echoes'} in video, and attribute the `less is more' paradox observed in Video-QA to this issue and the challenge of \textbf{`context dilution'}.
    \item We propose a novel, two-pronged framework that combines \textbf{\methodshort} with a lightweight, text-based semantic graph to act as a universal refinement module for any upstream keyframe selector.
    \item We conduct extensive experiments on the \lvb and \mme benchmarks across different baselines and models, demonstrating that our approach not only drastically reduces token costs but also frequently improves accuracy.
\end{itemize}

\section{Related Work}

\noindent\textbf{MLLMs for Video Understanding.}
The application of MLLMs to video understanding has seen rapid evolution through diverse paradigms. Early approaches utilized LLMs as high-level planners to generate executable code for video analysis~\cite{suris2023vipergpt, wu2023visual}. A dominant paradigm then emerged, focusing on direct visual-language alignment by connecting video encoders to LLMs via projection layers, enabling conversational abilities~\cite{maaz2024video, cheng2024videollama, zhang2023video, lin2024video, zhao2023learning, wu2025qwen}. To handle long-form videos, methods like MovieChat introduced memory mechanisms~\cite{song2024moviechat, weng2024longvlm}, while others combined visual features with textual signals like ASR/OCR for dense spatiotemporal tasks~\cite{yang2023vid2seq, luo2023valley}. More specialized adaptations have appeared, such as transforming LLMs into planners that iteratively call visual tools~\cite{wang2024videoagent}, regression models for timestamp prediction~\cite{huang2024vtimellm, ren2024timechat}, or efficient pipelines that leverage video analyzers to generate rich text descriptions for frozen LLMs~\cite{zhao2023learning}.

\noindent\textbf{Keyframe Selection for Video Understanding.}
A common thread uniting the aforementioned methods when applied to long videos is the prohibitive token cost. Consequently, keyframe selection has emerged as a crucial and dominant preprocessing paradigm~\cite{team2023gemini,hurst2024gpt,tang2025video}. Approaches to query-aware keyframe selection have diversified into several distinct families. One major family relies on \emph{clustering} frames based on visual or semantic similarity to find representative moments; for instance, VideoTree constructs a query-adaptive, hierarchical representation for coarse-to-fine reasoning~\cite{park2026too, mogrovejo2024question, zhang2024simple, wang2025videotree, jin2024chat}. Another popular paradigm frames the task as an \emph{agent-based} search, where the LLM iteratively interacts with the video to locate relevant information~\cite{fan2024videoagent, li2025videochat, wu2019adaframe, yu2023self}. A third, more recent school of thought focuses on direct \emph{temporal} search and optimization, developing sophisticated scoring mechanisms to pinpoint keyframes directly. State-of-the-art selectors like \basetstar~\cite{ye2025re}, \basevs~\cite{guologic}, and \baseaks~\cite{tang2025adaptive} exemplify this approach~\cite{liang2024keyvideollm}. Concurrently, retrieval-driven and learning-based techniques have gained traction. Moment Sampling leverages text-to-video retrieval models to isolate highly relevant temporal segments prior to frame extraction~\cite{chasmai2025moment}, whereas Frame-Voyager trains a dedicated querying module supervised by MLLM prediction losses to identify optimal frame combinations~\cite{yu2024frame}. Despite these advancements, existing keyframe selectors frequently prioritize high recall and suffer from two pitfalls: 1) generating redundant `visual echoes' that dilute context, and 2) the risk of losing semantic information during pruning. Our framework is designed to resolve both by using adaptive pruning to tackle redundancy and a semantic graph to compensate essential semantic information. 


\noindent\textbf{Addressing Redundancy and Semantic Gaps.}
Approaches to address temporal redundancy in video operate at various levels. While methods like TRIM~\cite{song2025less} perform patch-level token reduction within frames, our work is positioned at the frame level, consolidating entire redundant frames (``visual echoes''). To compensate for the potential semantic gap left by such pruning, a prominent paradigm is the use of structured semantic representations. A foundational approach in this area is the scene graph~\cite{krishna2017visual}, which provides a rich, structured vocabulary of objects and their inter-relationships. However, reasoning over these detailed graphs often necessitates complex Graph Neural Network (GNN) architectures~\cite{wu2020comprehensive}. More recent frameworks, such as CROSS~\cite{zhangunifying}, have explored modeling dynamic Temporal Text-attributed Graphs (TTAGs), demonstrating a trend towards integrating semantic and structural information over time. In contrast to these methods, our work introduces a lightweight text-attributed semantic graph specifically designed for minimal overhead. By injecting structured context directly into the MLLM's prompt, it serves as a highly compatible compensation mechanism without requiring specialized GNNs, aligning with maximum token efficiency.

\section{Methodology}

\begin{figure*}[t]
\centering
\includegraphics[width=\textwidth]{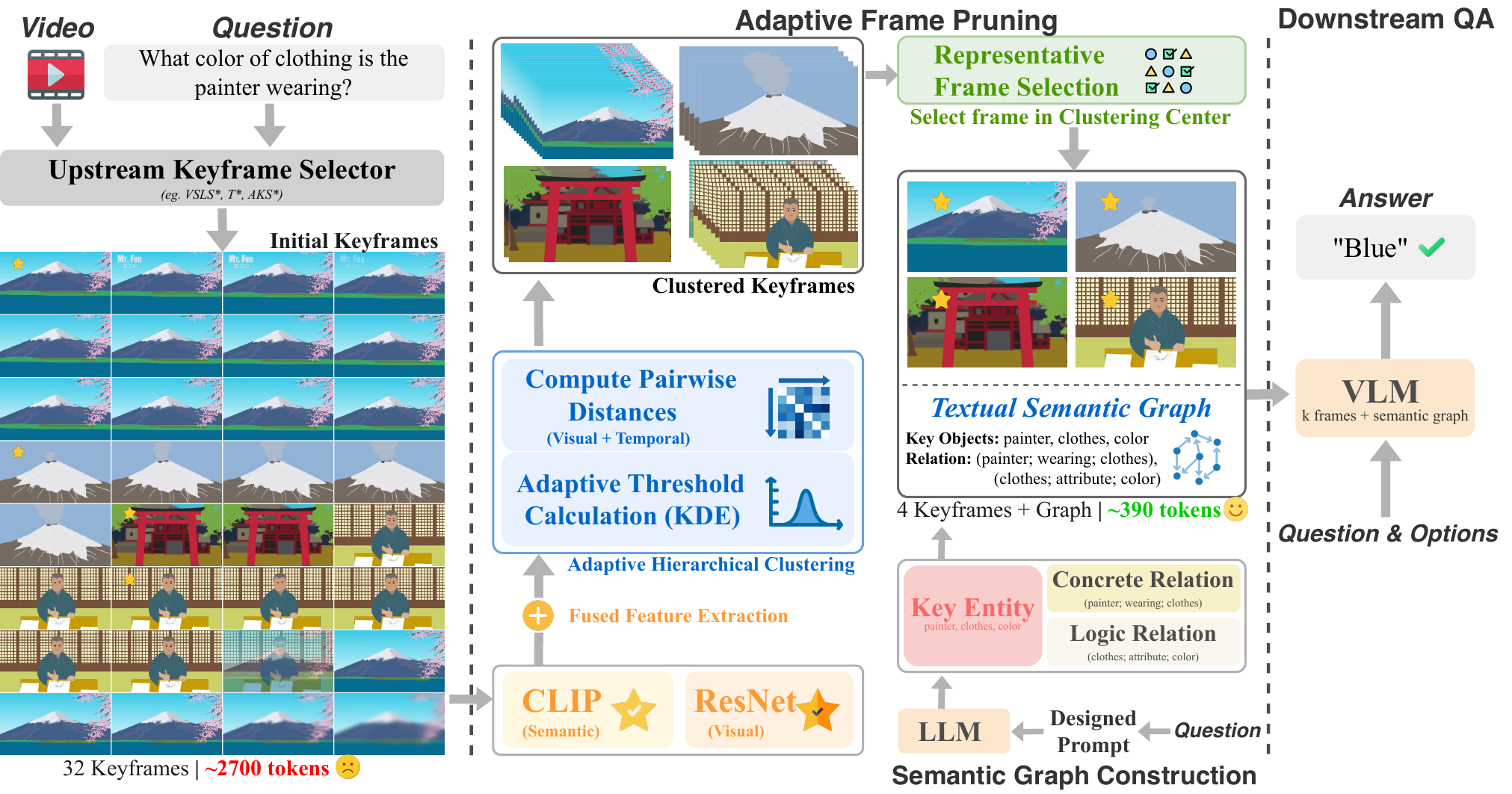}
\caption{\textbf{The Overall Pipeline of Our Proposed Method.} An upstream selector provides initial frames. Our \methodfullname module then takes over, performing (1) fused feature extraction and adaptive clustering to produce representative keyframes, and (2) concurrent semantic graph generation. Both are combined into an optimized prompt for MLLMs.}
\label{fig:pipeline}
\end{figure*}

Our approach enhances Video Understanding efficiency via a two-pronged strategy of \textbf{Pruning and Compensation}. We first prune the visual stream to a concise, non-redundant keyframe set, then compensate for information loss by injecting an efficient, abstract semantic layer. The two main stages, illustrated in Figure~\ref{fig:pipeline}, are: (1) \methodfullname (pruning) and (2) \textbf{Textual Semantic Graph Integration} (compensation).

\subsection{Adaptive Frame-Pruning (AFP)}
The core of our frame reduction strategy is \methodshort, an algorithm designed to identify and consolidate `visual echoes'. The process involves three key steps: extracting robust fused features, performing adaptive clustering, and selecting a representative frame from each cluster. The entire process is formalized in Algorithm~\ref{alg:afp}.

\subsubsection{Fused Feature Extraction.}
To ensure that our clustering is sensitive to both low-level visual patterns and high-level semantic content, we employ a fused feature representation. For each keyframe, we extract features from two powerful, pre-trained models: a ResNet-50~\cite{he2016deep} backbone, renowned for its strong performance on visual recognition tasks, and a CLIP ViT-B/32 model~\cite{radford2021learning}, which excels at aligning images with text and capturing rich semantic meaning. The high-dimensional outputs from both models are passed through separate linear projection layers to map them to a shared 512-dimensional space. The final feature vector $\mathbf{f}_{\text{fused}}$ for a frame is a weighted combination of the L2-normalized projected features:
\begin{equation}
    \mathbf{f}_{\text{fused}} = (1 - \alpha) \cdot \mathbf{f}_{\text{ResNet}} + \alpha \cdot \mathbf{f}_{\text{CLIP}},
    \label{eq:fusion}
\end{equation}
where $\alpha$ is a fusion ratio used to balance the contributions from the visual (ResNet) and semantic (CLIP) features.

\subsubsection{Adaptive Hierarchical Clustering.}
A key innovation of \methodshort is its ability to adapt the clustering process to the specific content of each video, avoiding a fixed, pre-defined number of clusters. We use Agglomerative Hierarchical Clustering~\cite{murtagh2012algorithms}, which requires a distance metric and a linkage criterion.

\noindent\textbf{Distance Metric.} Our distance metric jointly considers visual similarity and temporal proximity. For any two frames $i$ and $j$ with fused features $\mathbf{f}_i, \mathbf{f}_j$ and timestamps $t_i, t_j$, the combined distance $D(i, j)$ is defined as:
\begin{equation}
    D(i, j) = \beta \cdot d_{\text{cos}}( \mathbf{f}_i, \mathbf{f}_j ) + (1 - \beta) \cdot d_{\text{temp}}( t_i, t_j ),
    \label{eq:distance}
\end{equation}
where $d_{\text{cos}}$ is the cosine distance, $d_{\text{temp}}$ is the normalized absolute difference of timestamps, and $\beta$ is a weighting factor (set to 0.9). This combined metric ensures that frames that are visually similar but temporally distant are less likely to be clustered together.

\noindent\textbf{Adaptive Distance Threshold.} Instead of setting a fixed number of clusters, we use a dynamic distance threshold $\tau$. This threshold is determined by applying a Gaussian Kernel Density Estimator (KDE) to the distribution of all pairwise visual distances ($d_{\text{cos}}$) within the keyframe set. We identify the peak of this density distribution, which corresponds to the most common distance between frames, and set $\tau$ slightly above this value. This allows the algorithm to automatically find a natural grouping of frames based on their inherent similarity structure for that specific video.

\subsubsection{Representative Frame Selection.}
After clustering, a single representative frame must be selected from each resulting cluster. This is a critical step, as different selection strategies can impact performance. We conducted a detailed ablation study on several strategies (detailed in supplementary material). Our findings show that a \textbf{centroid-based selection} approach is the most robust and effective. Therefore, for all our main experiments, we identify the ``centroid frame'' for each cluster, the one with the minimum average visual feature distance to all other frames within that same cluster. This strategy selects the most visually representative frame, minimizing information loss during pruning without relying on external scores and thus enhancing the generalizability of our \methodshort framework.



\begin{algorithm}[tb]
\caption{\methodfullname}
\label{alg:afp}
\textbf{Input}: Initial keyframes $K = \{k_1, ..., k_N\}$, timestamps $T = \{t_1, ..., t_N\}$.\\
\textbf{Parameters}: Fusion ratio $\alpha$, distance weight $\beta$.\\
\textbf{Output}: Refined keyframes $K'$.

\begin{algorithmic}[1] 
\STATE Initialize feature set $F \leftarrow \emptyset$.
\FOR{each keyframe $k_i \in K$}
    \STATE Extract and project ResNet and CLIP features.
    \STATE Compute $\mathbf{f}_i$ using Eq.~\ref{eq:fusion}. 
    \STATE $F \leftarrow F \cup \{\mathbf{f}_i\}$.
\ENDFOR
\STATE Compute pairwise combined distance matrix $D$ using Eq.~\ref{eq:distance}.
\STATE Compute pairwise visual distances $D_{\text{cos}}$.
\STATE Apply KDE to $D_{\text{cos}}$ to find adaptive threshold $\tau$.
\STATE Perform Agglomerative Clustering on $K$ using distance matrix $D$ and threshold $\tau$ to get clusters $C = \{C_1, ..., C_M\}$.
\STATE Refine clusters by merging singleton clusters with their nearest neighbors.
\STATE Initialize refined set $K' \leftarrow \emptyset$.
\FOR{each cluster $C_j \in C$}
    \STATE Find $k^* = \arg\min_{k_i \in C_j} \sum_{k_m \in C_j} d_{\text{cos}}(\mathbf{f}_i, \mathbf{f}_m)$.
    \STATE $K' \leftarrow K' \cup \{k^*\}$.
\ENDFOR
\STATE \textbf{return} $K'$
\end{algorithmic}
\end{algorithm}

\subsection{Textual Semantic Graph Integration}


\noindent\textbf{Rationale and Generation.} 
The Textual Semantic Graph acts as a structured semantic scaffold to enhance MLLM reasoning, a principle validated by in-context learning methodologies~\cite{brown2020language, wei2022chain}. Generated by a lightweight, text-only LLM call that solely processes the query and options, this graph performs two critical functions: (1)~\textbf{Entity Grounding}, by identifying query-relevant objects and actors, and (2)~\textbf{Relationship Scaffolding}, by providing an explicit summary of their logical and concrete relations. Injecting this concise textual context primes the MLLM's reasoning pathways, enabling it to make more robust inferences from the pruned visual frames provided by \methodshort. Crucially, structuring this generation around explicit entity and relation extraction effectively mitigates the risk of textual hallucinations. A rigorous manual audit confirms a 98\% entity accuracy within our generated graphs (prompt and audit details are in the supplementary material).

\noindent\textbf{Token Efficiency.} 
The graph's utility is amplified by the asymmetric token costs between vision and text~\cite{OpenAIVisionGuide2024}. For instance, when refining a 32-frame selection from \baseaksstar on \lvb, our \methodshort module first reduces the visual token cost from approximately 2,877 to 505. The subsequent injection of a semantic graph, costing only about 60 tokens, introduces a negligible overhead. The immense savings from visual pruning thus vastly outweigh this cost, yielding a significant net reduction in tokens while boosting performance, as validated in our ablation studies (\cref{sec:ablation}). This design represents a deliberate and highly effective trade-off, where a minimal textual investment unlocks substantial gains in both efficiency and accuracy.

%
%
\section{Experiments}

\begin{table*}[t]
    \centering
    \setlength\tabcolsep{5pt} 
    \caption{\textbf{Main Results on \lvb and \mme with different MLLMs (GPT-4o, Qwen2.5-VL, and LLaVA-Video-7B).} We compare our full method (\methodgraphspace) against a sophisticated baseline (\baseaksstar) and a naive baseline (Uniform Sampling). Our method achieves superior or highly competitive accuracy while using drastically fewer frames (see Avg. Frames col.). \textbf{Bold} indicates the best performing method within each 3-row comparison block for each accuracy column.}
    \begin{adjustbox}{width=\textwidth}
    \begin{tabular}{l|c|ccc|l|c|ccc}
        \toprule
        \multicolumn{5}{c|}{\bf \lvb} & \multicolumn{5}{c}{\bf \mme} \\
        \cmidrule{1-10}
        \multirow{2}{*}{\bf Model and Method} & \bf Avg. & \multicolumn{3}{c|}{\textbf{Video Length Accuracy (\%)}} & \multirow{2}{*}{\bf Model and Method} & \bf Avg. & \multicolumn{3}{c}{\textbf{Video Length Accuracy (\%)}} \\
        & \bf Frames & Long & Medium & Short & & \bf Frames & Long & Medium & Short \\
        & & (900-3600s) & (180-600s) & (15-60s) & & & (30-60min) & (4-15min) & (0-2min) \\
        \bottomrule
        \multicolumn{10}{c}{\textit{Evaluation starting from Top 8 Keyframes based on \baseaksstar}} \\
        \hline
        GPT-4o + Uniform & 8.0 & 47.1 & 49.4 & 67.3 & GPT-4o + Uniform & 8.0 & \textbf{54.3} & \textbf{59.3} & \textbf{69.2} \\
        GPT-4o + \baseaksstar & 8.0 & 47.0 & 46.2 & 66.0 & GPT-4o + \baseaksstar & 8.0 & 51.6 & 52.7 & 55.8 \\
        \rowcolor{gray!10}
        GPT-4o + \methodgraph & \textbf{2.1} & \textbf{47.3} & \textbf{53.1} & \textbf{78.0} & GPT-4o + \methodgraph & \textbf{2.2} & 52.8 & 55.5 & 64.4 \\
        \hline 
        Qwen2.5-VL-7B + Uniform & 8.0 & 40.3 & 38.4 & 42.8 & Qwen2.5-VL-7B + Uniform & 8.0 & 36.6 & \textbf{39.1} & 41.1 \\
        Qwen2.5-VL-7B + \baseaksstar & 8.0 & 35.4 & 41.5 & 56.0 & Qwen2.5-VL-7B + \baseaksstar & 8.0 & \textbf{39.4} & 37.8 & 42.2 \\
        \rowcolor{gray!10}
        Qwen2.5-VL-7B + \methodgraph & \textbf{2.1} & \textbf{42.9} & \textbf{46.5} & \textbf{64.0} & Qwen2.5-VL-7B + \methodgraph & \textbf{2.1} & 37.0 & 37.1 & \textbf{48.9} \\
        \hline
        LLaVA-Video-7B + Uniform & 8.0 & 40.2 & 46.9 & 50.0 & LLaVA-Video-7B + Uniform & 8.0 & 37.4 & 41.1 & 41.4 \\
        LLaVA-Video-7B + \baseaksstar & 8.0 & 41.4 & 45.8 & 52.0 & LLaVA-Video-7B + \baseaksstar & 8.0 & 38.0 & 42.2 & 40.5 \\
        \rowcolor{gray!10}
        LLaVA-Video-7B + \methodgraph & \textbf{2.1} & \textbf{44.6} & \textbf{53.1} & \textbf{62.0} & LLaVA-Video-7B + \methodgraph & \textbf{2.1} & \textbf{43.9} & \textbf{46.0} & \textbf{55.2} \\
        \bottomrule
        \multicolumn{10}{c}{\textit{Evaluation starting from Top 32 Keyframes based on \baseaksstar}} \\
        \hline
        GPT-4o + Uniform & 32.0 & \textbf{50.6} & 53.5 & 74.0 & GPT-4o + Uniform & 32.0 & 55.2 & \textbf{61.0} & \textbf{71.4} \\
        GPT-4o + \baseaksstar & 32.0 & 47.0 & 49.2 & 58.0 & GPT-4o + \baseaksstar & 32.0 & \textbf{55.6} & 54.5 & 57.8 \\
        \rowcolor{gray!10}
        GPT-4o + \methodgraph & \textbf{4.1} & 49.1 & \textbf{53.5} & \textbf{84.0} & GPT-4o + \methodgraph & \textbf{4.2} & 55.0 & 56.7 & 65.4 \\
        \hline
        Qwen2.5-VL-7B + Uniform & 32.0 & 32.7 & 36.5 & 50.0 & Qwen2.5-VL-7B + Uniform & 32.0 & 33.6 & 38.1 & \textbf{60.0} \\
        Qwen2.5-VL-7B + \baseaksstar & 32.0 & 35.7 & 42.7 & 60.0 & Qwen2.5-VL-7B + \baseaksstar & 32.0 & 37.4 & 39.7 & 43.8 \\
        \rowcolor{gray!10}
        Qwen2.5-VL-7B + \methodgraph & \textbf{4.1} & \textbf{43.5} & \textbf{46.9} & \textbf{64.0} & Qwen2.5-VL-7B + \methodgraph & \textbf{4.2} & \textbf{39.2} & \textbf{41.3} & 50.9 \\
        \hline
        LLaVA-Video-7B + Uniform & 32.0 & 41.7 & 46.2 & 40.0 & LLaVA-Video-7B + Uniform & 32.0 & 39.1 & 41.7 & 39.9 \\
        LLaVA-Video-7B + \baseaksstar & 32.0 & 43.5 & 46.5 & 50.0 & LLaVA-Video-7B + \baseaksstar & 32.0 & 39.6 & 40.9 & 41.6 \\
        \rowcolor{gray!10}
        LLaVA-Video-7B + \methodgraph & \textbf{4.1} & \textbf{45.2} & \textbf{53.8} & \textbf{64.0} & LLaVA-Video-7B + \methodgraph & \textbf{4.2} & \textbf{42.6} & \textbf{46.6} & \textbf{53.9} \\
        \bottomrule
    \end{tabular}
    \end{adjustbox}
\label{tab:main_results_final}
\end{table*}

\subsection{Experimental Setup}
\label{sec:exp_setup}

\noindent\textbf{Datasets and Models.} 
We evaluate our method on two challenging long-video question-answering benchmarks: \textbf{\lvb}~\cite{wu2024longvideobench} and \textbf{\mme}~\cite{fu2025video}. To demonstrate the generalizability, we conduct experiments across three distinct MLLMs: the proprietary \textbf{GPT-4o}~\cite{hurst2024gpt}, and two powerful open-source models, \textbf{Qwen2.5-VL-7B-Instruct}~\cite{bai2025qwen2} and \textbf{LLaVA-Video-7B-Qwen2}~\cite{li2024llava}.

\noindent\textbf{Compared Methods and Baselines.}
In our main experiments, we use \baseaksstar as our primary upstream selector to demonstrate the core effectiveness of our method. Our evaluation protocol is designed to rigorously assess our method. For each MLLM, we compare two keyframe sampling strategies in our main results (Table~\ref{tab:main_results_final}):
\begin{itemize}
    \item \textbf{Uniform Sampling}: A naive baseline where a fixed number of frames (32 or 8) are sampled at regular intervals.
    \item \textbf{\baseaksstar:} An adaptation of the state of the art \baseaks~\cite{tang2025adaptive} framework. To ensure a fair comparison under a constant frame budget, we replace its original dynamic sampling with Top K selection based on its internal BLIP relevance scores. This variant, denoted as \baseaksstar, follows the ablation protocol established in the original work and ensures a robust, reproducible baseline (see supplementary material).
    \item \methodgraphspace: Our full method, which takes the output of the upstream selector as input and applies our adaptive pruning and semantic graph compensation.
\end{itemize} 
To further validate the generalizability of our approach, we also apply it to two other state-of-the-art selectors, \basetstar and \basevsstar. The results of these extensive generalizability studies are provided in supplementary material.




\noindent\textbf{Evaluation Metrics.} We evaluate all methods based on three primary criteria:
\begin{itemize}
    \item \textbf{Effectiveness:} Official \textbf{Accuracy (\%)} calculated via strict exact matching of the predicted option letter against the ground truth.
    \item \textbf{Efficiency:} \textbf{Average Frames} per query, defined as $\text{AvgFrames} = \frac{1}{N} \sum_{i=1}^{N} F_i$, where $F_i$ represents the frame count for the $i$-th instance.
    \item \textbf{Token Efficiency:} \textbf{Estimated token cost} for GPT-4o, formulated as $\overline{T} = \frac{1}{N} \sum_{i=1}^{N} (T^{\text{text}}_i + 85 \cdot F_i)$. Here, $T^{\text{text}}_i$ is the textual prompt token count, $F_i$ denotes the number of low-detail images, and 85 is the per-image cost prescribed by OpenAI~\cite{OpenAIVisionGuide2024}.
\end{itemize}


\noindent\textbf{Implementation and Hyperparameters.}
For our main experiments, the core hyperparameters were set to $\alpha=0.6$ for the feature fusion ratio and $\beta=0.9$ for the distance metric weight. These values were determined through a systematic sensitivity analysis, which revealed a clear trade-off between performance and efficiency. A detailed analysis of these hyperparameters and further implementation details are provided in supplementary material.

\subsection{Main Results and Analysis}



\noindent\textbf{The ``Less is More'' Paradox.} Before analyzing our method's performance, we empirically validate the core challenge our framework targets. Table~\ref{tab:main_results_final} reveals that on \lvb with GPT-4o (8-frame budget), the sophisticated \baseaksstar selector unexpectedly underperforms the naive Uniform Sampling baseline (47.0\% versus 47.1\%). This confirms that maximizing recall without selective filtering inundates MLLMs with `visual echoes', leading to severe `context dilution' that degrades reasoning capabilities. This finding dictates that a dedicated refinement stage is essential to unlock the true potential of query-aware selection.

\noindent\textbf{Superior Performance and Efficiency.} Table~\ref{tab:main_results_final} demonstrates that our complete \methodgraphspace framework unequivocally surpasses baseline methods. The most profound impact is observed on open-source models: using LLaVA-Video-7B on \lvb (32-frame budget), our method boosts accuracy on short videos from 50.0\% (\baseaksstar) to \textbf{64.0\%} using merely 4.1 frames on average. This 14-point improvement indicates that open-source models are highly susceptible to context dilution and benefit immensely from our focused prompts. Furthermore, our approach effectively rescues struggling baselines. For the \lvb GPT-4o (8-frame) case, our framework applied to \baseaksstar output achieves 53.1\% on Medium videos, surpassing both baselines while utilizing 74\% fewer frames.

\begin{figure}[t]
\centering
\includegraphics[width=\columnwidth]{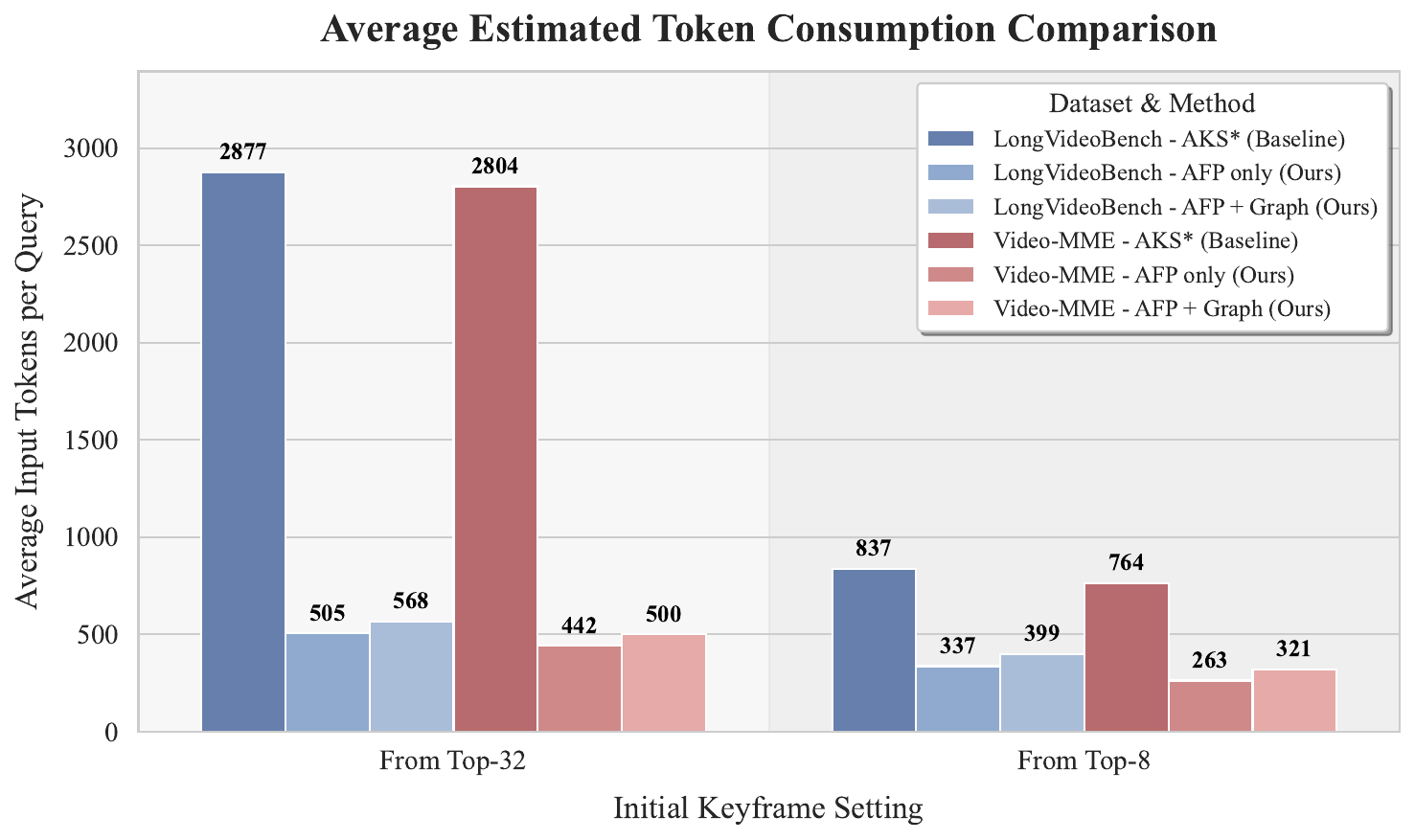}
\caption{\textbf{Average estimated token consumption comparison across datasets and methods.} Token counts are estimated based on OpenAI's guidelines (see Section~\ref{sec:exp_setup} for details). Our method (\methodgraphspace) reduces the token requirements compared to the \baseaksstar baseline across all settings and on both datasets.
}
\label{fig:token_comparison}
\end{figure}

\begin{figure}[t]
\centering
\includegraphics[width=\columnwidth]{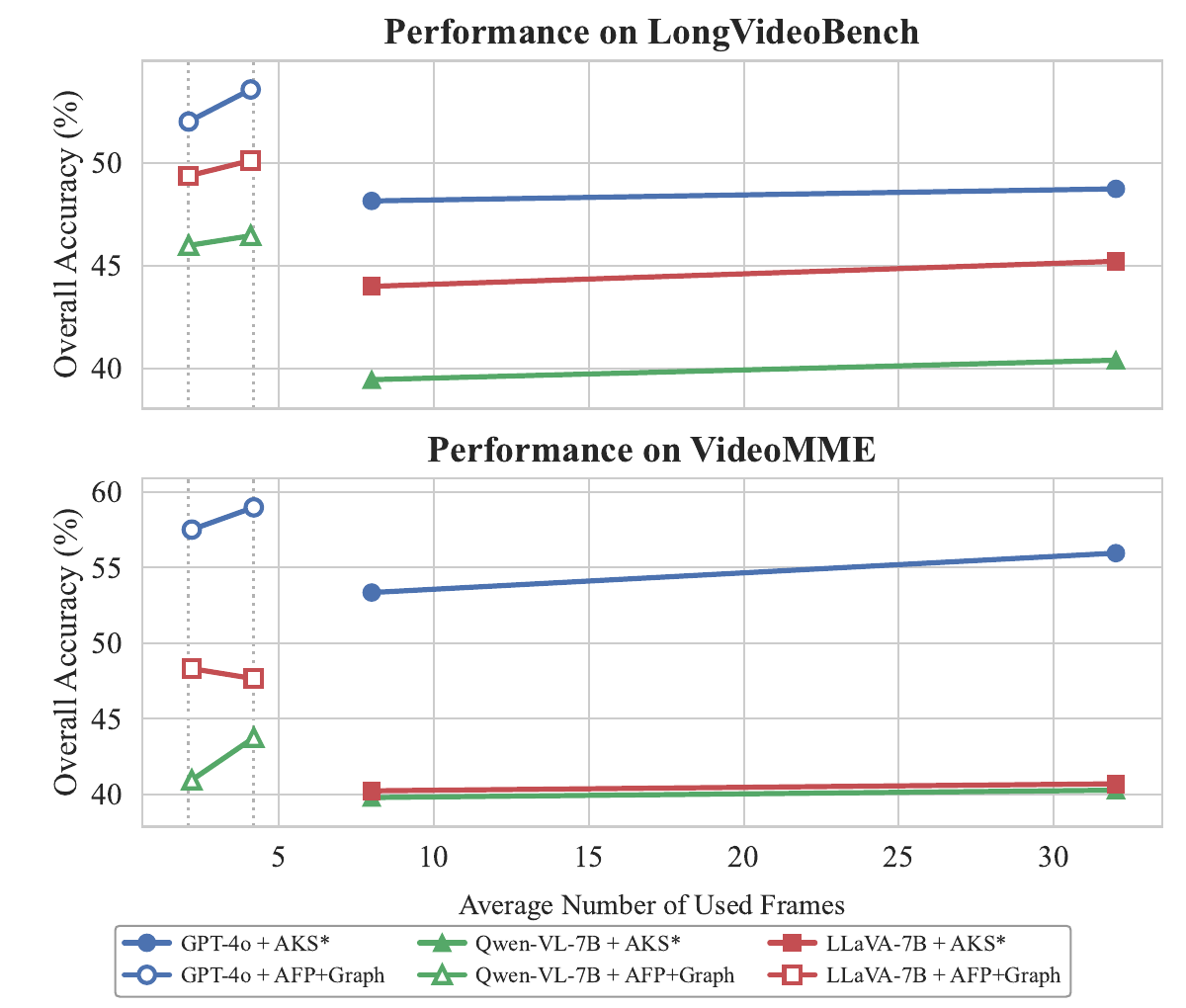}
\caption{\textbf{Efficiency and Performance results of \methodgraphspace on \baseaksstar Keyframes across different MLLMs.} The plot shows the trade-off between overall accuracy and the average number of frames used on both datasets. 
}
\label{fig:main_plot_vsls}
\end{figure}

\noindent\textbf{Token Reduction and Universal Applicability.} This immense efficiency gain is visualized in Figure~\ref{fig:token_comparison} and Figure~\ref{fig:main_plot_vsls}. On \lvb (Top-32), our method slashes the estimated token count from 2,877 (\baseaksstar) to just 568, an over \textbf{5$\times$ reduction}. Figure~\ref{fig:main_plot_vsls} confirms our method consistently operates in the optimal top-left quadrant, achieving a SOTA efficiency-performance trade-off. Importantly, our framework is inherently agnostic to the upstream selector. As detailed in the supplementary material, integrating our refinement layer with other leading methods (\basevsstar, \basetstar) consistently yields substantial gains in most scenarios, with particularly strong improvements on short videos. For instance, refining \basetstar on the \mme dataset (Top-32) boosts short video accuracy by 13.4 absolute points using an average of 4.1 frames, confirming our method's versatility.

\noindent\textbf{Robustness on Event Dense Videos.}
We also conducted a stress test on the highly dynamic Video-HOLMES~\cite{cheng2025video} benchmark. As detailed in the supplementary material, our method remains robust in such challenging scenarios, achieving a 3.47\% accuracy improvement over the baseline.

\subsection{Quantifying Visual Redundancy Mitigation}

\begin{table}[h!]
\centering
\footnotesize
\setlength{\tabcolsep}{3pt} 
\renewcommand{\arraystretch}{1.1} 
\caption{\textbf{Quantitative analysis of `visual echoes' mitigation.} We report the Average Maximum Similarity (AMS) and average frame count (Fr.) before (Base) and after applying our framework (Ours) to Top-32 keyframes. Lower AMS values indicate less redundancy.}
\begin{tabularx}{\columnwidth}{@{} l c c | c c @{}}
    \toprule
    \multirow{2}{*}{\textbf{Upstream Selector}} & \multicolumn{2}{c|}{\textbf{LONGVIDEOBENCH}} & \multicolumn{2}{c}{\textbf{VIDEOMME}} \\
    \cmidrule(lr){2-3} \cmidrule(lr){4-5}
    & \textbf{AMS} (Base$\to$Ours) & \textbf{Fr.} & \textbf{AMS} (Base$\to$Ours) & \textbf{Fr.} \\
    \midrule
    \baseaksstar & 0.950 $\rightarrow$ \textbf{0.826} & 4.1 & 0.936 $\rightarrow$ \textbf{0.802} & 4.2 \\
    \basevsstar  & 0.932 $\rightarrow$ \textbf{0.776} & 4.2 & 0.929 $\rightarrow$ \textbf{0.742} & 4.3 \\
    \basetstar    & 0.937 $\rightarrow$ \textbf{0.796} & 4.2 & 0.934 $\rightarrow$ \textbf{0.795} & 4.1 \\
    \bottomrule
\end{tabularx}
\label{tab:ams_quantification}
\end{table}

To quantify redundancy, we introduce the \textbf{Average Maximum Similarity (AMS)} metric, where higher values denote greater inter-frame visual similarity. The analysis presented in Table~\ref{tab:ams_quantification} reveals a stark contrast. While baseline selectors consistently yield highly redundant outputs with AMS scores exceeding 0.93, our framework systematically reduces this redundancy, achieving significantly more diverse sets with AMS scores ranging from 0.74 to 0.83. This result validates that ``visual echoes'' are a prevalent artifact of existing methods and that our approach provides an effective mitigation.

\begin{figure}[h!] 
\centering
\includegraphics[width=\columnwidth]{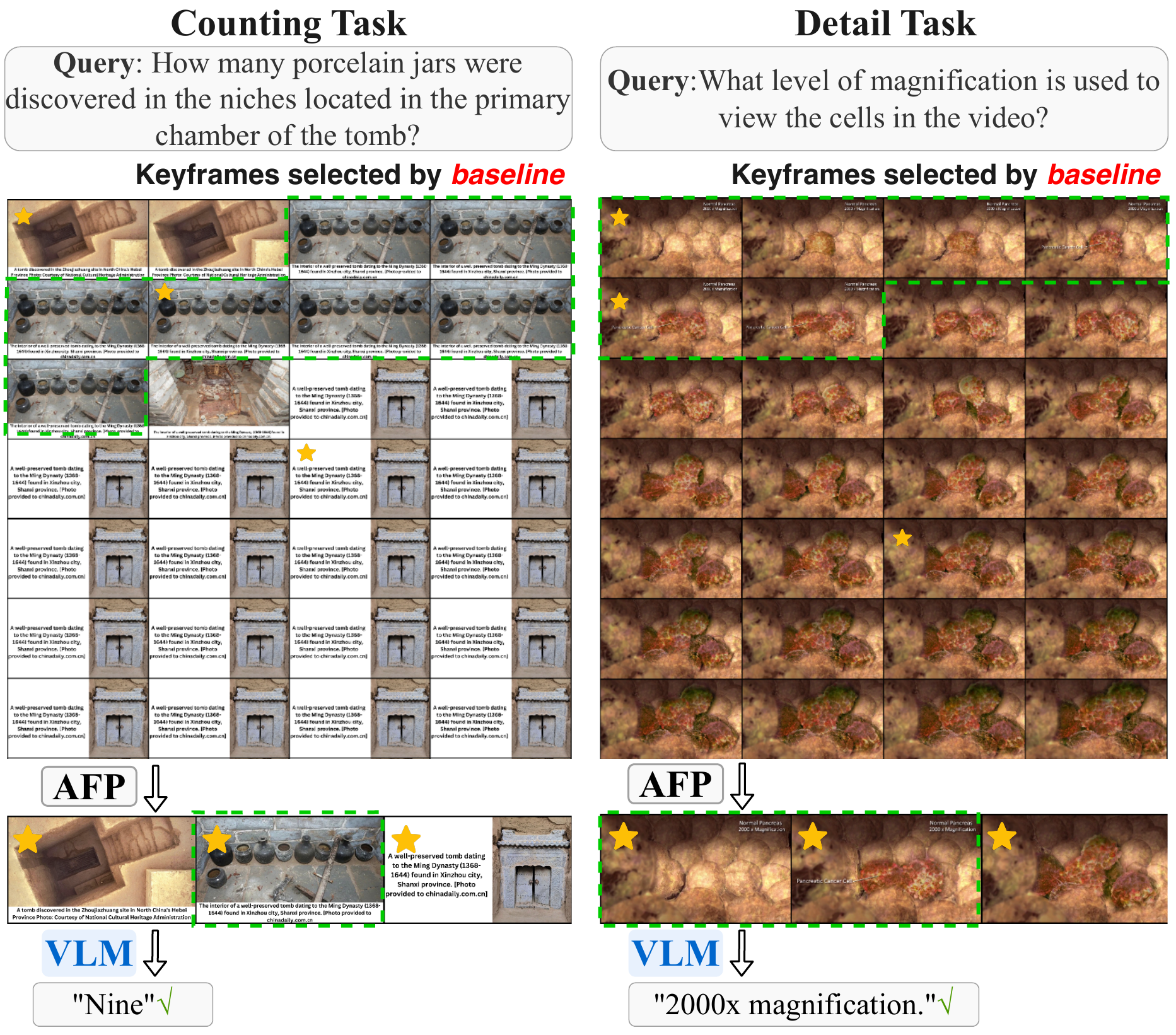} 
\caption{\textbf{Qualitative demonstration of redundancy pruning.} We present a counting task (left) and a fine detail recognition task (right). In both scenarios, the baseline method selects numerous repetitive views (highlighted by green dashed boxes). Our \methodshort algorithm intelligently consolidates these `visual echoes', pruning the initial sets to just 3 highly informative frames.}
\label{fig:qualitative_pruning_main}
\end{figure}

This mechanism is visually demonstrated in Figure~\ref{fig:qualitative_pruning_main}. We showcase two challenging scenarios: a counting task requiring global object aggregation and a detail recognition task requiring localized text identification. In both instances, the baseline selector retrieves thirty two frames laden with highly overlapping views of the target subjects. Our \methodshort algorithm correctly identifies these repetitive frames as `visual echoes' and merges them into a highly compact set of merely three representative frames. Critically, this intelligent consolidation retains the holistic visual evidence required for aggregation (left) while safeguarding the semantically decisive local details for recognition (right), demonstrating a robust balance between effectiveness and efficiency.

\begin{table*}[t!]
\centering
\small 
\setlength{\tabcolsep}{3.5pt} 
\caption{\textbf{Comprehensive component-wise ablation study under a matched frame budget.} We compare our full method against baselines including text-only, graph-only, uniform sampling, and \baseaksstar truncation. Our full method consistently outperforms all others.}
\begin{minipage}[t]{0.49\textwidth}
    \centering
    \begin{adjustbox}{width=\linewidth}
    \begin{tabular}{l|c|c|ccc}
    \toprule
    \textbf{Method} & \textbf{Avg. Fr.} & \textbf{Avg. Tok.} & \textbf{Long(\%)} & \textbf{Med.(\%)} & \textbf{Short(\%)} \\
    \bottomrule
    \multicolumn{6}{c}{\textit{Ablation from \baseaksstar Top 32 Keyframes (Model: GPT-4o)}} \\
    \hline
    Text Only & 0.0 & 157.05 & 42.9 & 45.8 & 48.0 \\
    Graph Only & 0.0 & 219.48 & 47.0 & 53.5 & 82.0 \\
    Uniform (M) & 4.1 & 505.21 & 47.0 & 47.7 & 66.0 \\
    \baseaksstar (Top-$N$, M) & 4.1 & 505.21 & 43.5 & 45.8 & 62.0 \\
    \methodshort only & 4.1 & 505.21 & 47.9 & 45.8 & 64.0 \\
    \methodgraphspace & \textbf{4.1} & \textbf{567.64} & \textbf{49.1} & \textbf{53.5} & \textbf{84.0} \\
    \bottomrule
    \multicolumn{6}{c}{\textit{Ablation from \baseaksstar Top 8 Keyframes (Model: Qwen2.5-VL-7B)}} \\
    \hline
    Text Only & 0.0 & N/A & 35.1 & 38.1 & 52.0 \\
    Graph Only & 0.0 & N/A & 37.5 & 41.2 & 56.0 \\
    Uniform (M) & 2.1 & N/A & 39.0 & 38.8 & 62.0 \\
    \baseaksstar (Top-$N$, M) & 2.1 & N/A & 41.4 & 41.5 & 56.0 \\
    \methodshort only & 2.1 & N/A & 41.4 & 42.3 & 60.0 \\
    \methodgraphspace & \textbf{2.1} & N/A & \textbf{42.9} & \textbf{46.5} & \textbf{64.0} \\
    \bottomrule
    \end{tabular}
    \end{adjustbox}
    \subcaption{Results on the \lvb dataset.}
    \label{tab:ablation_lvb}
\end{minipage}
\hfill 
\begin{minipage}[t]{0.49\textwidth}
    \centering
    \begin{adjustbox}{width=\linewidth}
    \begin{tabular}{l|c|c|ccc}
    \toprule
    \textbf{Method} & \textbf{Avg. Fr.} & \textbf{Avg. Tok.} & \textbf{Long(\%)} & \textbf{Med.(\%)} & \textbf{Short(\%)} \\
    \bottomrule
    \multicolumn{6}{c}{\textit{Ablation from \baseaksstar Top 32 Keyframes (Model: GPT-4o)}} \\
    \hline
    Text Only & 0.0 & 84.10 & 43.9 & 42.7 & 35.4 \\
    Graph Only & 0.0 & 141.52 & 51.1 & 54.3 & 60.4 \\
    Uniform (M) & 4.2 & 442.18 & 53.1 & 53.3 & 59.2 \\
    \baseaksstar (Top-$N$, M) & 4.2 & 442.18 & 51.1 & 50.0 & 53.2 \\
    \methodshort only & 4.2 & 442.18 & 53.7 & 51.6 & 57.1 \\
    \methodgraphspace & \textbf{4.2} & \textbf{499.60} & \textbf{55.0} & \textbf{56.7} & \textbf{65.4} \\
    \bottomrule
    \multicolumn{6}{c}{\textit{Ablation from \baseaksstar Top 8 Keyframes (Model: LLaVA-Video-7B)}} \\
    \hline
    Text Only & 0.0 & N/A & 37.4 & 40.4 & 41.6 \\
    Graph Only & 0.0 & N/A & 42.3 & 46.0 & 54.2 \\
    Uniform (M) & 2.1 & N/A & 39.2 & 40.6 & 39.7 \\
    \baseaksstar (Top-$N$, M) & 2.1 & N/A & 37.8 & 40.8 & 41.1 \\
    \methodshort only & 2.1 & N/A & 40.6 & 40.7 & 40.7 \\
    \methodgraphspace & \textbf{2.1} & N/A & \textbf{43.9} & \textbf{46.0} & \textbf{55.2} \\
    \bottomrule
    \end{tabular}
    \end{adjustbox}
    \subcaption{Results on the \mme dataset.}
    \label{tab:ablation_mme}
\end{minipage}
\label{tab:comprehensive_ablation}
\end{table*}

\section{Ablation Studies and Experiment Analysis}
\label{sec:ablation}

\subsection{Component-Level Ablation Analysis}
To isolate the individual contributions of our core modules, Table~\ref{tab:comprehensive_ablation} presents a component-wise ablation study. Token counts are estimated as detailed in Section~\ref{sec:exp_setup}. For a rigorous and fair evaluation, all visual strategies operate under a strictly matched frame budget dynamically determined by our \methodshort algorithm for each specific Video-QA instance. We compare our complete framework against five baselines:


\begin{itemize}
    \item \textbf{Text Only:} Establishes a non-visual lower bound using solely the question and options. 
    \item \textbf{Graph Only:} Evaluates the standalone reasoning power of our textual semantic graph.
    \item \textbf{Uniform (Matched):} Implements naive sampling restricted to the target frame count.
    \item \textbf{\baseaksstar (Top-$N$, Matched):} Truncates the \baseaksstar ranked list to meet the exact budget.
    \item \textbf{\methodshort only:} Removes the semantic graph to isolate the efficacy of intelligent visual clustering.
    \item \textbf{\methodgraphspace:} Our complete refinement framework.
\end{itemize}

\begin{figure}[t!]
    \centering
    \includegraphics[width=1.0\linewidth]{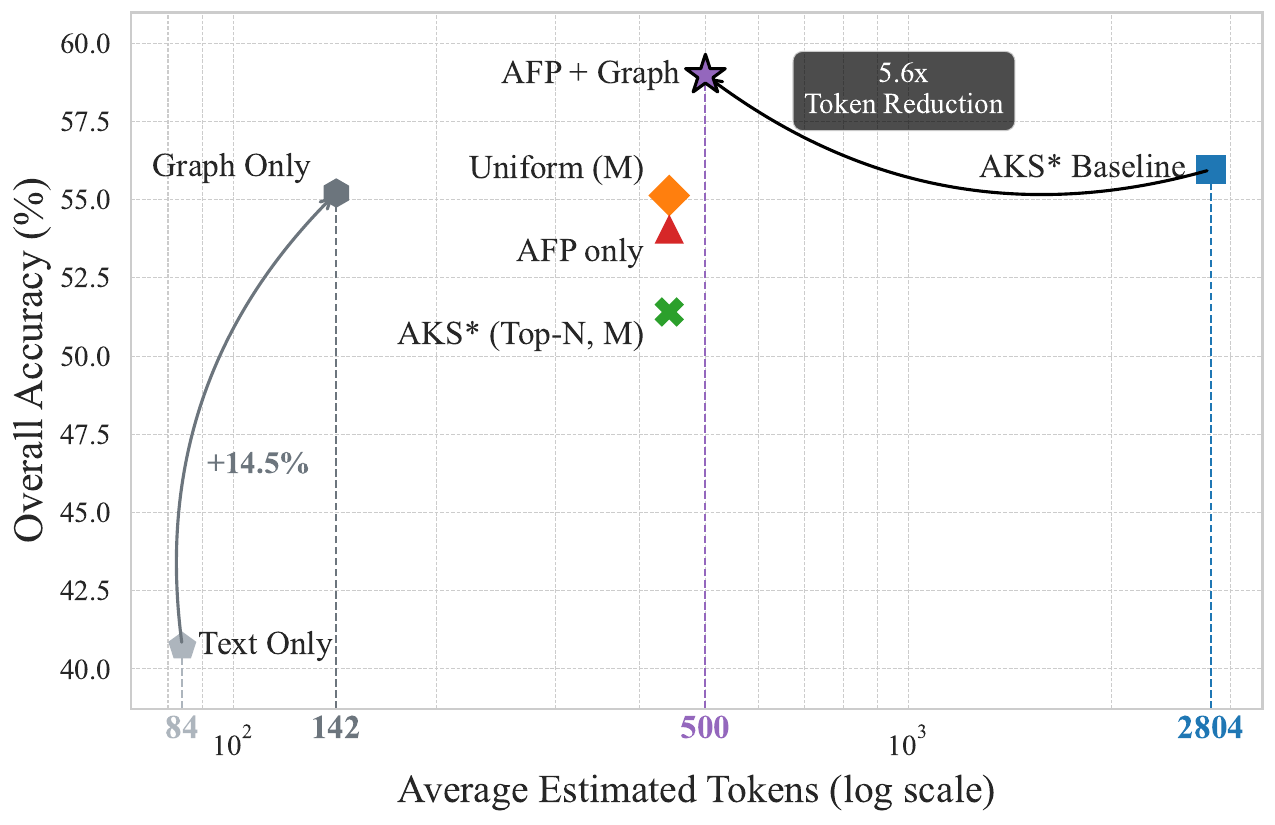} 
    \caption{\textbf{Efficiency-Performance Trade-off Analysis.} This scatter plot visualizes the results from our ablation study on \mme based on the \baseaksstar Top-32 setting with GPT-4o. The x-axis represents the average token cost (log scale), while the y-axis shows overall QA accuracy. \methodgraphspace achieved the highest accuracy with a 5.6x token reduction compared with \baseaksstar.}
    \label{fig:ablation_tradeoff}
\end{figure}

\noindent\textbf{Synergy of Vision and Semantics.}
The results highlight the immense standalone potency of the semantic graph. As illustrated in Figure~\ref{fig:ablation_tradeoff}, the `Graph Only' approach yields a +14.5\% absolute accuracy gain over the `Text Only' baseline. This indicates that our LLM-based graph construction functions as a highly effective form of knowledge distillation. However, visual evidence remains irreplaceable. In every setting, \methodgraphspace consistently outperforms the `Graph Only' configuration. For instance, on \lvb (Short) using Qwen2.5, adding just 2.1 frames boosts accuracy from 56.0\% to 64.0\%. This confirms that while the semantic graph provides a vital reasoning scaffold, the concise visual cues selected by \methodshort are indispensable.

\noindent\textbf{Efficacy of Adaptive Pruning.}
When evaluating purely frame-based strategies, the superiority of \methodshort is evident. For GPT-4o on \lvb (Long), `\methodshort only' (47.9\%) outperforms both the `Uniform (Matched)' (47.0\%) and the sophisticated `\baseaksstar (Top-$N$, Matched)' (43.5\%) baselines. Figure~\ref{fig:ablation_tradeoff} visually encapsulates our overarching contribution: our complete \methodgraphspace framework shifts performance into the optimal top-left quadrant. It achieves peak overall accuracy while drastically cutting token costs by \textbf{5.6$\times$} compared to the \baseaksstar baseline. This empirically validates our core premise that refining verbose visual inputs into token-efficient, semantically potent prompts directly enhances reasoning capabilities.


\subsection{End-to-End Inference Efficiency}

To evaluate the practical viability of our framework, we measured the end-to-end wall-clock inference time, encompassing all preprocessing, model inference, and network latencies. The results, presented in Table~\ref{tab:inference_efficiency}, reveal a significant acceleration across all tested configurations. This efficiency gain stems from alleviating critical inference bottlenecks: our method reduces network I/O overhead for API-based models and mitigates the substantial computational cost of visual token pre-filling for locally-deployed models. The minimal overhead introduced by our refinement stage is vastly outweighed by these savings, confirming the framework's practical value in real-world applications.

\begin{table}[h!]
\centering
\footnotesize
\setlength{\tabcolsep}{4pt} 
\renewcommand{\arraystretch}{1.1} 
\caption{\textbf{End-to-end inference latency analysis.} We report the total wall-clock time (minutes) and the speedup ratio (Spd.) comparing the baseline (Base) against our full framework (Ours). Our method accelerates inference for both API-based and local models.}
\begin{tabularx}{\columnwidth}{@{} l c c | c c @{}}
    \toprule
    \multirow{2}{*}{\textbf{Model}} & \multicolumn{2}{c|}{\textbf{LONGVIDEOBENCH}} & \multicolumn{2}{c}{\textbf{VIDEOMME}} \\
    \cmidrule(lr){2-3} \cmidrule(lr){4-5}
    & \textbf{Time} (Base$\to$Ours) & \textbf{Spd.} & \textbf{Time} (Base$\to$Ours) & \textbf{Spd.} \\
    \midrule
    GPT-4o & 106m $\rightarrow$ \textbf{68m} & \textbf{1.6$\times$} & 463m $\rightarrow$ \textbf{275m} & \textbf{1.7$\times$} \\
    Qwen2.5 & 143m $\rightarrow$ \textbf{104m} & \textbf{1.4$\times$} & 664m $\rightarrow$ \textbf{301m} & \textbf{2.2$\times$} \\
    \bottomrule
\end{tabularx}
\label{tab:inference_efficiency}
\vspace{-4mm} 
\end{table}
\section{Conclusion}

We addressed the prohibitive token costs in Video-QA by demonstrating that concise inputs yield superior multimodal reasoning. To mitigate the severe temporal redundancy formalized as \textbf{`visual echoes'}, we proposed a synergistic framework comprising \methodfullname and a \textbf{textual semantic graph}. Extensive evaluations confirm that our refinement strategy, by pruning visual redundancy and injecting a semantic scaffold, effectively filters context noise, drastically reducing computational overhead while elevating accuracy. Future research will extend this paradigm toward a `Multimodal Semantic Graph' integrating audio cues and motion trajectories. This overcomes the global feature limitations detailed in the supplementary material, delivering a comprehensive yet compact video representation to enhance reasoning stability over extensive sequences.

\section*{Acknowledgements}

This research was supported by the High-Performance Computing (HPC) resources at the Hong Kong University of Science and Technology (Guangzhou) and the GPU cluster of the research group led by Prof. Hui Xiong. The computational costs associated with the GPT API were generously funded by the College of Future Technology at HKUST(GZ). We express our sincere gratitude to our advisors, Prof. Xuming Hu and Prof. Hui Xiong, for their invaluable guidance and mentorship throughout this work.

{
    \small
    \bibliographystyle{ieeenat_fullname}
    \bibliography{main} 
}

\appendix 

\clearpage
%
%
\section{Generalizability Studies on \basevsstar}
\label{sec:ablation_vs}

\paragraph{A Note on \basevsstar.}
In our generalizability study, we use a specific, publicly available version of the \basevs~\cite{guologic} codebase from May 2025, which we denote as \basevsstar. It is important to note that this version differs slightly from the final published version of \basevs. The primary difference lies in the sampling strategy: this earlier version normalizes the relevance scores into a probability distribution before sampling, whereas the final version performs a direct Top-K selection on the raw scores. We use this specific, frozen version to ensure the reproducibility of our experiments presented here.

\paragraph{Experimental Setup.}
This section provides the complete component-wise ablation results for applying our framework to the output of \basevsstar. 

The methods compared are:
\begin{itemize}
    \item \textbf{\basevsstar Baseline}: The two high-cost methods using a fixed number of frames (32 or 8) via `\basevsstar'.
    \item \textbf{Matched-Budget Strategies}: Four low-cost methods that all operate on the same, drastically reduced number of frames determined by our \methodshort algorithm for each Video-QA instance.
    \begin{itemize}
        \item Uniform (Matched): Naive uniform sampling.
        \item \basevsstar (Top-N, Matched): A strong baseline that truncates the \basevsstar list.
        \item \methodshort only (Ours): Our pruning method without the graph.
        \item \methodgraphspace (Ours): Our full proposed method.
    \end{itemize}
\end{itemize}
The detailed results are presented in Table~\ref{tab:full_ablation_appendix_vsls}.

\paragraph{Analysis from Tabular Data.}
The results in Table~\ref{tab:full_ablation_appendix_vsls} robustly confirm that our framework acts as a versatile refinement module for \basevsstar.
First, our full \methodgraphspace{} method achieves a highly favorable efficiency-performance trade-off. The performance gains are particularly striking on open-source models. For instance, on \lvb with LLaVA-Video-7B (from 8 frames), our method boosts the accuracy on short videos from 50.0\% (\basevsstar Baseline) to a remarkable \textbf{72.0\%}, a \textbf{+22 point} absolute improvement—while using only 2.2 frames.
Second, the superiority of our frame selection is evident. Under a strictly matched frame budget, `\methodshort~only' (e.g., on \mme with GPT-4o from 32 frames) demonstrates clear performance advantages over `\basevsstar (Top-N, Matched)', indicating that our intelligent clustering selects a more representative set of frames.
In addition, the semantic graph proves its crucial role. The consistent and significant performance leap from `\methodshort~only' to `\methodgraphspace' across all settings validates our rationale that the graph provides the essential semantic scaffolding to unlock the MLLM's full potential.

\begin{figure}[h!]
    \centering
    \includegraphics[width=\columnwidth]{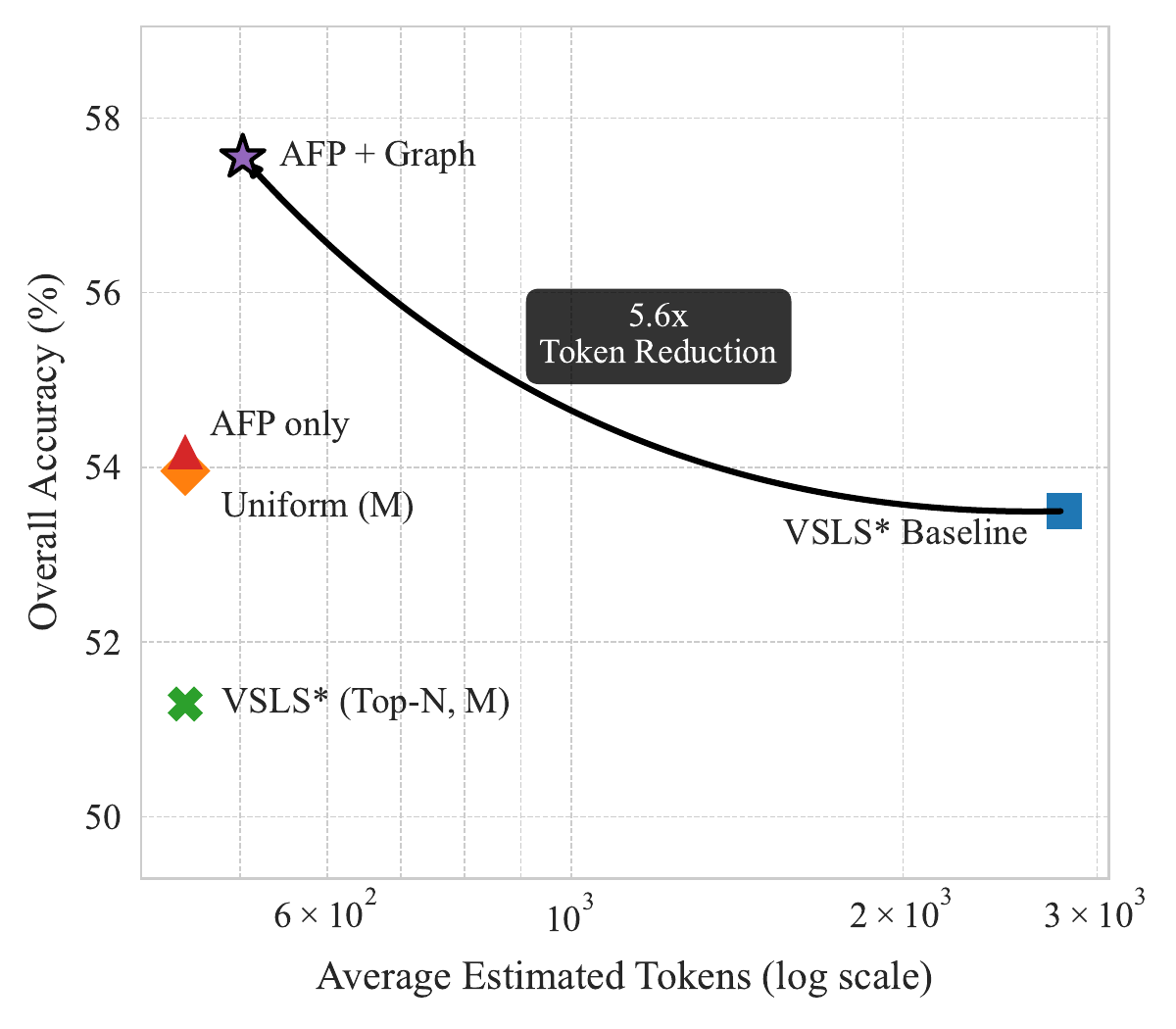}
    \caption{\textbf{Efficiency-Performance Trade-off on \mme with \basevsstar.} This plot visualizes the results for GPT-4o when refining the Top-32 output from \basevsstar. The x-axis (log scale) represents token cost, and the y-axis represents weighted average accuracy. The top-left region is optimal.}
    \label{fig:appendix_scatter_vsls}
\end{figure}

\paragraph{Visual Analysis of Efficiency-Performance.}
This trade-off is best visualized in the efficiency-performance plot in Figure~\ref{fig:appendix_scatter_vsls}, which focuses on the challenging \textbf{GPT-4o on \mme (from Top-32)} scenario. The plot clearly illustrates \methodgraphspace's superiority. The \basevsstar Baseline resides in the bottom-right, representing high cost and moderate performance. Within the highly competitive low-cost zone on the left, our full \methodgraphspace method distinguishes itself, positioned highest in the desirable top-left quadrant. It not only significantly outperforms the `\basevsstar (Top-N, Matched)' baseline (57.55\% vs. 51.29\%) but also surpasses the `\methodshort only' approach (54.18\%), demonstrating the crucial performance boost provided by the semantic graph.

\begin{table*}[t!]
\centering
\small 
\setlength{\tabcolsep}{4pt} 
\caption{\textbf{Complete component-wise ablation study results on the \lvb and \mme datasets using \basevsstar as the upstream selector.} For each model and initial frame setting, we compare our full method against various baselines. \textbf{Bold} indicates the best performance among the four matched-budget strategies for each accuracy column.}
\begin{minipage}[t]{0.49\textwidth}
    \centering
    \begin{adjustbox}{width=\linewidth}
    \begin{tabular}{l|c|ccc}
    \toprule
    \textbf{Method} & \textbf{Avg. Fr.} & \textbf{Long(\%)} & \textbf{Med.(\%)} & \textbf{Short(\%)} \\
    \bottomrule
    \multicolumn{5}{c}{\textit{Evaluation from Top 8 Keyframes}} \\
    \hline
    \multicolumn{5}{l}{\textbf{Model: GPT-4o}} \\
    \basevsstar Baseline & 8.0 & 44.6 & 46.9 & 66.0 \\
    Uniform (Matched) & 2.2 & 47.6 & 48.1 & 62.0 \\
    \basevsstar (Top-N, Matched) & 2.2 & \textbf{50.3} & 48.1 & 66.0 \\
    \methodshort only (Ours) & 2.2 & 45.5 & 51.5 & 56.0 \\
    \methodgraphspace (Ours) & \textbf{2.2} & 47.3 & \textbf{53.5} & \textbf{84.0} \\
    \hline
    \multicolumn{5}{l}{\textbf{Model: Qwen2.5-VL-7B-Instruct}} \\
    \basevsstar Baseline & 8.0 & \textbf{45.8} & \textbf{49.2} & 54.0 \\
    Uniform (Matched) & 2.2 & 38.1 & 41.9 & \textbf{70.0} \\
    \basevsstar (Top-N, Matched) & 2.2 & 44.0 & 47.3 & 56.0 \\
    \methodshort only (Ours) & 2.2 & 40.2 & 45.0 & 68.0 \\
    \methodgraphspace (Ours) & \textbf{2.2} & 42.6 & 45.4 & 62.0 \\
    \hline
    \multicolumn{5}{l}{\textbf{Model: LLaVA-Video-7B-Qwen2}} \\
    \basevsstar Baseline & 8.0 & 42.6 & 43.5 & 50.0 \\
    Uniform (Matched) & 2.2 & 39.3 & 46.2 & 42.0 \\
    \basevsstar (Top-N, Matched) & 2.2 & 40.5 & 42.7 & 46.0 \\
    \methodshort only (Ours) & 2.2 & 41.1 & 44.6 & 44.0 \\
    \methodgraphspace (Ours) & \textbf{2.2} & \textbf{45.5} & \textbf{49.6} & \textbf{72.0} \\
    \bottomrule
    \multicolumn{5}{c}{\textit{Evaluation from Top 32 Keyframes}} \\
    \hline
    \multicolumn{5}{l}{\textbf{Model: GPT-4o}} \\
    \basevsstar Baseline & 32.0 & 46.1 & 45.0 & 76.0 \\
    Uniform (Matched) & 4.2 & 47.9 & 48.8 & 64.0 \\
    \basevsstar (Top-N, Matched) & 4.2 & \textbf{50.9} & \textbf{53.8} & 64.0 \\
    \methodshort only (Ours) & 4.2 & 45.5 & 47.7 & 66.0 \\
    \methodgraphspace (Ours) & \textbf{4.2} & 49.4 & 51.5 & \textbf{80.0} \\
    \hline
    \multicolumn{5}{l}{\textbf{Model: Qwen2.5-VL-7B-Instruct}} \\
    \basevsstar Baseline & 32.0 & 38.7 & 42.3 & 54.0 \\
    Uniform (Matched) & 4.2 & 45.8 & 45.0 & 68.0 \\
    \basevsstar (Top-N, Matched) & 4.2 & \textbf{50.3} & \textbf{53.1} & 66.0 \\
    \methodshort only (Ours) & 4.2 & 40.2 & 41.9 & 60.0 \\
    \methodgraphspace (Ours) & \textbf{4.2} & 42.9 & 46.9 & \textbf{66.0} \\
    \hline
    \multicolumn{5}{l}{\textbf{Model: LLaVA-Video-7B-Qwen2}} \\
    \basevsstar Baseline & 32.0 & 41.7 & 48.1 & 54.0 \\
    Uniform (Matched) & 4.2 & 38.1 & 47.3 & 52.0 \\
    \basevsstar (Top-N, Matched) & 4.2 & 40.2 & 45.8 & 52.0 \\
    \methodshort only (Ours) & 4.2 & 41.7 & 43.5 & 50.0 \\
    \methodgraphspace (Ours) & \textbf{4.2} & \textbf{45.2} & \textbf{50.0} & \textbf{62.0} \\
    \bottomrule
    \end{tabular}
    \end{adjustbox}
    \subcaption{Results on the \lvb dataset.}
    \label{tab:full_ablation_appendix_lvb}
\end{minipage}
\hfill 
\begin{minipage}[t]{0.49\textwidth}
    \centering
    \begin{adjustbox}{width=\linewidth}
    \begin{tabular}{l|c|ccc}
    \toprule
    \textbf{Method} & \textbf{Avg. Fr.} & \textbf{Long(\%)} & \textbf{Med.(\%)} & \textbf{Short(\%)} \\
    \bottomrule
    \multicolumn{5}{c}{\textit{Evaluation from Top 8 Keyframes}} \\
    \hline
    \multicolumn{5}{l}{\textbf{Model: GPT-4o}} \\
    \basevsstar Baseline & 8.0 & 51.7 & 52.4 & 56.5 \\
    Uniform (Matched) & 2.1 & 52.4 & 53.4 & 58.6 \\
    \basevsstar (Top-N, Matched) & 2.1 & 50.5 & 49.8 & 56.0 \\
    \methodshort only (Ours) & 2.1 & 53.1 & 51.8 & 57.3 \\
    \methodgraphspace (Ours) & \textbf{2.1} & \textbf{53.5} & \textbf{56.5} & \textbf{63.8} \\
    \hline
    \multicolumn{5}{l}{\textbf{Model: Qwen2.5-VL-7B-Instruct}} \\
    \basevsstar Baseline & 8.0 & 36.6 & 39.1 & 41.1 \\
    Uniform (Matched) & 2.1 & 38.5 & 40.2 & 42.9 \\
    \basevsstar (Top-N, Matched) & 2.1 & 38.1 & 37.9 & 41.4 \\
    \methodshort only (Ours) & 2.1 & 39.2 & 38.8 & 42.4 \\
    \methodgraphspace (Ours) & \textbf{2.1} & \textbf{38.1} & \textbf{40.2} & \textbf{50.1} \\
    \hline
    \multicolumn{5}{l}{\textbf{Model: LLaVA-Video-7B-Qwen2}} \\
    \basevsstar Baseline & 8.0 & 38.7 & 41.5 & 41.3 \\
    Uniform (Matched) & 2.1 & 40.6 & 41.2 & 38.3 \\
    \basevsstar (Top-N, Matched) & 2.1 & 39.2 & 40.2 & 38.5 \\
    \methodshort only (Ours) & 2.1 & 38.9 & 41.0 & 40.3 \\
    \methodgraphspace (Ours) & \textbf{2.1} & \textbf{44.2} & \textbf{47.9} & \textbf{54.5} \\
    \bottomrule
    \multicolumn{5}{c}{\textit{Evaluation from Top 32 Keyframes}} \\
    \hline
    \multicolumn{5}{l}{\textbf{Model: GPT-4o}} \\
    \basevsstar Baseline & 32.0 & 51.7 & 52.4 & 56.5 \\
    Uniform (Matched) & 4.3 & 52.3 & 52.6 & 57.1 \\
    \basevsstar (Top-N, Matched) & 4.3 & 49.7 & 50.2 & 54.1 \\
    \methodshort only (Ours) & 4.3 & 52.5 & 52.5 & 57.7 \\
    \methodgraphspace (Ours) & \textbf{4.3} & \textbf{53.0} & \textbf{56.4} & \textbf{63.4} \\
    \hline
    \multicolumn{5}{l}{\textbf{Model: Qwen2.5-VL-7B-Instruct}} \\
    \basevsstar Baseline & 32.0 & 37.9 & 39.1 & \textbf{55.8} \\
    Uniform (Matched) & 4.3 & 39.1 & 43.5 & 43.6 \\
    \basevsstar (Top-N, Matched) & 4.3 & 37.9 & 40.2 & 42.5 \\
    \methodshort only (Ours) & 4.3 & 38.3 & 39.8 & 42.5 \\
    \methodgraphspace (Ours) & \textbf{4.3} & \textbf{39.9} & \textbf{44.0} & 51.8 \\
    \hline
    \multicolumn{5}{l}{\textbf{Model: LLaVA-Video-7B-Qwen2}} \\
    \basevsstar Baseline & 32.0 & 37.4 & 40.6 & 42.1 \\
    Uniform (Matched) & 4.3 & 39.6 & 40.6 & 41.7 \\
    \basevsstar (Top-N, Matched) & 4.3 & 38.1 & 41.1 & 41.2 \\
    \methodshort only (Ours) & 4.3 & 39.8 & 40.9 & 41.5 \\
    \methodgraphspace (Ours) & \textbf{4.3} & \textbf{43.8} & \textbf{47.9} & \textbf{56.7} \\
    \bottomrule
    \end{tabular}
    \end{adjustbox}
    \subcaption{Results on the \mme dataset.}
    \label{tab:full_ablation_appendix_mme}
\end{minipage}
\label{tab:full_ablation_appendix_vsls}
\end{table*}

%
%
\section{Generalizability Studies on \basetstar}
\label{sec:ablation_tstar}

\paragraph{Experimental Setup.}
To further validate the universal applicability of our framework, we conducted a comprehensive set of experiments applying it to another SOTA keyframe selector, \basetstar~\cite{ye2025re}. The comparison follows the same structure as our \basevsstar study, with detailed results presented in Table~\ref{tab:full_ablation_appendix_tstar}.

\paragraph{Analysis from Tabular Data.}
The results in Table~\ref{tab:full_ablation_appendix_tstar} further solidify our framework's universal applicability. Our full method consistently offers a superior cost-utility trade-off, with gains being most pronounced on open-source models. For instance, when refining the \basetstar output for LLaVA-Video-7B on \mme (from 32 frames), our method elevates the accuracy on short videos from 41.6\% to \textbf{55.0\%}, a \textbf{+13.4 point} improvement. The quality of frames selected by `\methodshort~only' is demonstrably high. Under the matched-budget setting for GPT-4o on \lvb (from 8 frames), `\methodshort~only' (47.3\% on Long videos) outperforms both `Uniform (Matched)' (45.5\%) and `\basetstar (Top-N, Matched)' (46.4\%). In addition, the semantic graph provides a significant performance boost, reinforcing our central thesis about the synergy between our two components.

\begin{figure}[h!]
    \centering
    \includegraphics[width=\columnwidth]{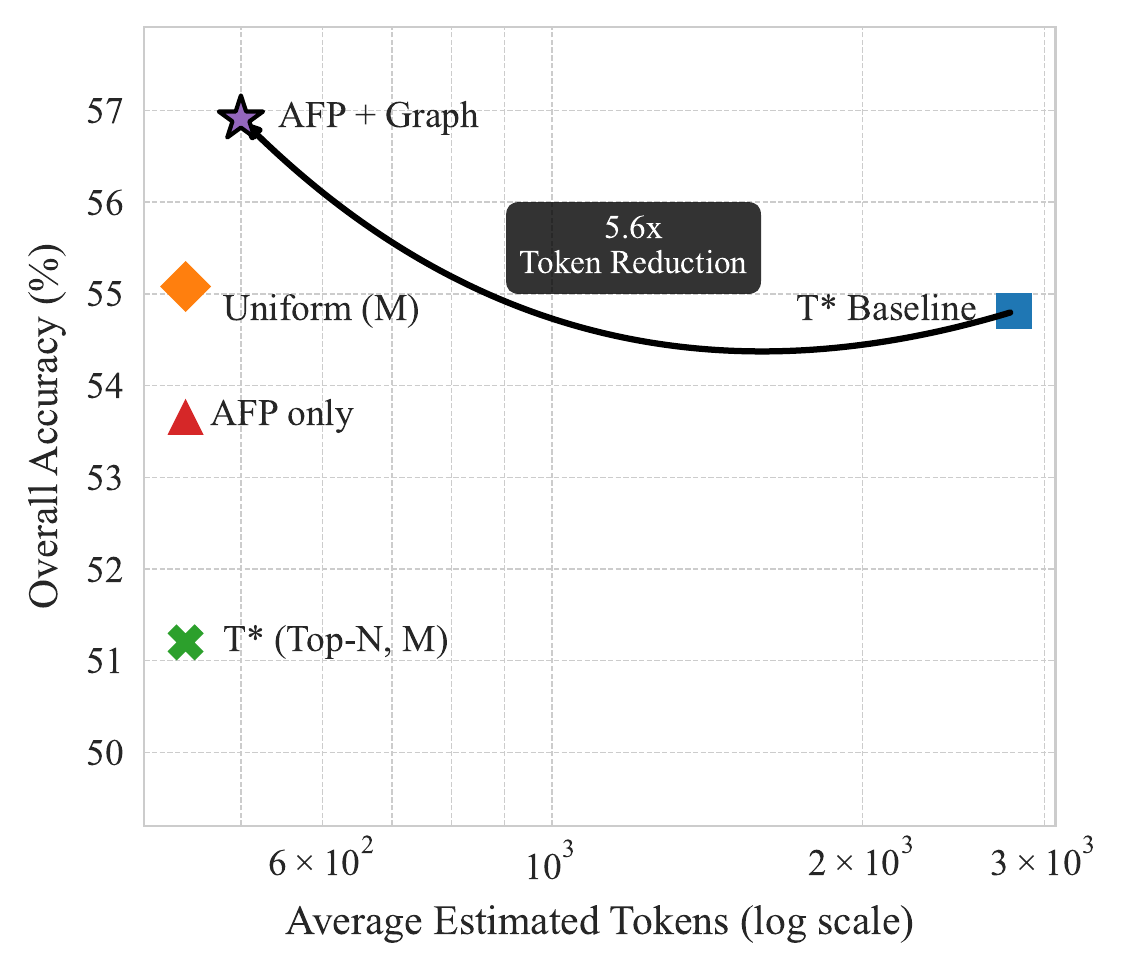}
    \caption{\textbf{Efficiency-Performance Trade-off on \mme with \basetstar.} This plot visualizes the results for GPT-4o when refining the Top-32 output from \basetstar. The top-left region is optimal.}
    \label{fig:appendix_scatter_tstar}
\end{figure}

\paragraph{Visual Analysis of Efficiency-Performance.}
This trade-off is powerfully visualized in the efficiency-performance plot in Figure~\ref{fig:appendix_scatter_tstar}, which highlights the GPT-4o on \mme (from Top-32) scenario. Similar to the \basevsstar case, the \basetstar resides in the high-cost, bottom-right region. Within the highly competitive low-cost zone, our full \methodgraphspace method once again establishes itself as the superior choice, occupying the highest point in the desirable top-left quadrant. It significantly outperforms the `\basetstar (Top-N, Matched)' baseline (56.91\% vs. 51.20\%) and shows a clear improvement over the `\methodshort only' approach (53.66\%). This parallel success story proves that our framework's effectiveness is not tied to a specific selector but stems from its fundamental principles of intelligent pruning and semantic compensation.

\begin{table*}[t!]
\centering
\small 
\setlength{\tabcolsep}{4pt} 
\caption{\textbf{Generalizability study on the \basetstar selector.} This table shows the complete component-wise ablation results of applying our framework to the output of \basetstar on both datasets. \textbf{Bold} indicates the best performance among the four matched-budget strategies for each accuracy column.}
\begin{minipage}[t]{0.49\textwidth}
    \centering
    \begin{adjustbox}{width=\linewidth}
    \begin{tabular}{l|c|ccc}
    \toprule
    \textbf{Method} & \textbf{Avg. Fr.} & \textbf{Long(\%)} & \textbf{Med.(\%)} & \textbf{Short(\%)} \\
    \bottomrule
    \multicolumn{5}{c}{\textit{Evaluation from Top 8 Keyframes}} \\
    \hline
    \multicolumn{5}{l}{\textbf{Model: GPT-4o}} \\
    \basetstar Baseline & 8.0 & 44.3 & 46.2 & 66.0 \\
    Uniform (Matched) & 2.2 & 45.5 & 48.8 & 64.0 \\
    \basetstar (Top-N, Matched) & 2.2 & 46.4 & \textbf{53.1} & 64.0 \\
    \methodshort only (Ours) & 2.2 & \textbf{47.3} & 48.8 & 68.0 \\
    \methodgraphspace (Ours) & \textbf{2.2} & 46.4 & 52.7 & \textbf{80.0} \\
    \hline
    \multicolumn{5}{l}{\textbf{Model: Qwen2.5-VL-7B-Instruct}} \\
    \basetstar Baseline & 8.0 & 42.0 & 47.7 & 54.0 \\
    Uniform (Matched) & 2.2 & 35.7 & 38.5 & \textbf{70.0} \\
    \basetstar (Top-N, Matched) & 2.2 & \textbf{43.8} & \textbf{48.1} & 56.0 \\
    \methodshort only (Ours) & 2.2 & 38.7 & 41.9 & 66.0 \\
    \methodgraphspace (Ours) & \textbf{2.2} & 42.0 & 45.4 & 66.0 \\
    \hline
    \multicolumn{5}{l}{\textbf{Model: LLaVA-Video-7B-Qwen2}} \\
    \basetstar Baseline & 8.0 & 42.0 & 47.7 & 44.0 \\ 
    Uniform (Matched) & 2.2 & 40.8 & 46.5 & 48.0 \\
    \basetstar (Top-N, Matched) & 2.2 & 39.3 & 45.4 & 42.0 \\
    \methodshort only (Ours) & 2.2 & 39.6 & 49.2 & 52.0 \\
    \methodgraphspace (Ours) & \textbf{2.2} & \textbf{45.5} & \textbf{51.5} & \textbf{62.0} \\
    \bottomrule
    \multicolumn{5}{c}{\textit{Evaluation from Top 32 Keyframes}} \\
    \hline
    \multicolumn{5}{l}{\textbf{Model: GPT-4o}} \\
    \basetstar Baseline & 32.0 & 53.1 & 48.8 & 74.3 \\
    Uniform (Matched) & 4.2 & 46.4 & 48.8 & 62.0 \\
    \basetstar (Top-N, Matched) & 4.2 & \textbf{49.1} & \textbf{55.0} & 62.0 \\
    \methodshort only (Ours) & 4.2 & 45.8 & 48.8 & 70.0 \\
    \methodgraphspace (Ours) & \textbf{4.2} & 48.5 & 53.1 & \textbf{80.0} \\
    \hline
    \multicolumn{5}{l}{\textbf{Model: Qwen2.5-VL-7B-Instruct}} \\
    \basetstar Baseline & 32.0 & 38.7 & 41.9 & 40.0 \\
    Uniform (Matched) & 4.2 & 38.1 & 43.5 & 64.0 \\
    \basetstar (Top-N, Matched) & 4.2 & 41.4 & \textbf{48.5} & \textbf{68.0} \\
    \methodshort only (Ours) & 4.2 & 41.1 & 41.2 & 62.0 \\
    \methodgraphspace (Ours) & \textbf{4.2} & \textbf{42.6} & 43.1 & \textbf{68.0} \\
    \hline
    \multicolumn{5}{l}{\textbf{Model: LLaVA-Video-7B-Qwen2}} \\
    \basetstar Baseline & 32.0 & 40.2 & 44.2 & 50.0 \\
    Uniform (Matched) & 4.2 & 38.7 & 46.9 & 48.0 \\
    \basetstar (Top-N, Matched) & 4.2 & \textbf{42.3} & 46.9 & 48.0 \\
    \methodshort only (Ours) & 4.2 & 41.1 & 49.2 & 44.0 \\
    \methodgraphspace (Ours) & \textbf{4.2} & 42.0 & \textbf{52.7} & \textbf{62.0} \\
    \bottomrule
    \end{tabular}
    \end{adjustbox}
    \subcaption{Results on the \lvb dataset.}
    \label{tab:tstar_ablation_lvb}
\end{minipage}
\hfill 
\begin{minipage}[t]{0.49\textwidth}
    \centering
    \begin{adjustbox}{width=\linewidth}
    \begin{tabular}{l|c|ccc}
    \toprule
    \textbf{Method} & \textbf{Avg. Fr.} & \textbf{Long(\%)} & \textbf{Med.(\%)} & \textbf{Short(\%)} \\
    \bottomrule
    \multicolumn{5}{c}{\textit{Evaluation from Top 8 Keyframes}} \\
    \hline
    \multicolumn{5}{l}{\textbf{Model: GPT-4o}} \\
    \basetstar Baseline & 8.0 & 51.3 & 51.5 & 55.2 \\
    Uniform (Matched) & 2.1 & 50.1 & 54.2 & 58.6 \\
    \basetstar (Top-N, Matched) & 2.1 & 49.9 & 51.7 & 56.2 \\
    \methodshort only (Ours) & 2.1 & 52.2 & 51.1 & 56.7 \\
    \methodgraphspace (Ours) & \textbf{2.1} & \textbf{52.3} & \textbf{55.3} & \textbf{66.1} \\
    \hline
    \multicolumn{5}{l}{\textbf{Model: Qwen2.5-VL-7B-Instruct}} \\
    \basetstar Baseline & 8.0 & \textbf{38.2} & 38.3 & 43.1 \\
    Uniform (Matched) & 2.1 & 37.7 & 39.4 & 42.5 \\
    \basetstar (Top-N, Matched) & 2.1 & 36.1 & 38.5 & 41.4 \\
    \methodshort only (Ours) & 2.1 & 38.0 & 39.3 & 41.5 \\
    \methodgraphspace (Ours) & \textbf{2.1} & 36.2 & \textbf{40.4} & \textbf{49.5} \\
    \hline
    \multicolumn{5}{l}{\textbf{Model: LLaVA-Video-7B-Qwen2}} \\
    \basetstar Baseline & 8.0 & 40.4 & 41.3 & 41.6 \\
    Uniform (Matched) & 2.1 & 38.8 & 41.6 & 39.2 \\
    \basetstar (Top-N, Matched) & 2.1 & 39.3 & 41.0 & 38.8 \\
    \methodshort only (Ours) & 2.1 & 37.8 & 40.4 & 40.0 \\
    \methodgraphspace (Ours) & \textbf{2.1} & \textbf{42.3} & \textbf{46.2} & \textbf{56.1} \\
    \bottomrule
    \multicolumn{5}{c}{\textit{Evaluation from Top 32 Keyframes}} \\
    \hline
    \multicolumn{5}{l}{\textbf{Model: GPT-4o}} \\
    \basetstar Baseline & 32.0 & 52.6 & 54.4 & 57.5 \\
    Uniform (Matched) & 4.1 & 52.4 & 55.2 & 57.7 \\
    \basetstar (Top-N, Matched) & 4.1 & 49.8 & 49.4 & 54.6 \\
    \methodshort only (Ours) & 4.1 & 51.8 & 53.6 & 55.7 \\
    \methodgraphspace (Ours) & \textbf{4.1} & \textbf{52.6} & \textbf{55.3} & \textbf{63.1} \\
    \hline
    \multicolumn{5}{l}{\textbf{Model: Qwen2.5-VL-7B-Instruct}} \\
    \basetstar Baseline & 32.0 & 36.9 & 40.2 & 42.2 \\
    Uniform (Matched) & 4.1 & \textbf{39.9} & \textbf{43.1} & 43.5 \\
    \basetstar (Top-N, Matched) & 4.1 & 37.9 & 40.1 & 43.4 \\
    \methodshort only (Ours) & 4.1 & 39.6 & 40.4 & 44.6 \\
    \methodgraphspace (Ours) & \textbf{4.1} & 39.4 & 40.7 & \textbf{52.3} \\
    \hline
    \multicolumn{5}{l}{\textbf{Model: LLaVA-Video-7B-Qwen2}} \\
    \basetstar Baseline & 32.0 & 38.2 & 42.2 & 41.6 \\
    Uniform (Matched) & 4.1 & 38.7 & 42.9 & 41.3 \\
    \basetstar (Top-N, Matched) & 4.1 & 40.2 & 39.3 & 40.3 \\
    \methodshort only (Ours) & 4.1 & 38.4 & 41.7 & 39.3 \\
    \methodgraphspace (Ours) & \textbf{4.1} & \textbf{41.7} & \textbf{47.1} & \textbf{55.0} \\
    \bottomrule
    \end{tabular}
    \end{adjustbox}
    \subcaption{Results on the \mme dataset.}
    \label{tab:tstar_ablation_mme}
\end{minipage}
\label{tab:full_ablation_appendix_tstar}
\end{table*}

%
%
\section{Additional Visual Evidence for the Prevalence of `visual echoes'}
\label{sec:appendix_additional_echoes}

\begin{figure}[h!]
    \centering
    \includegraphics[width=\columnwidth]{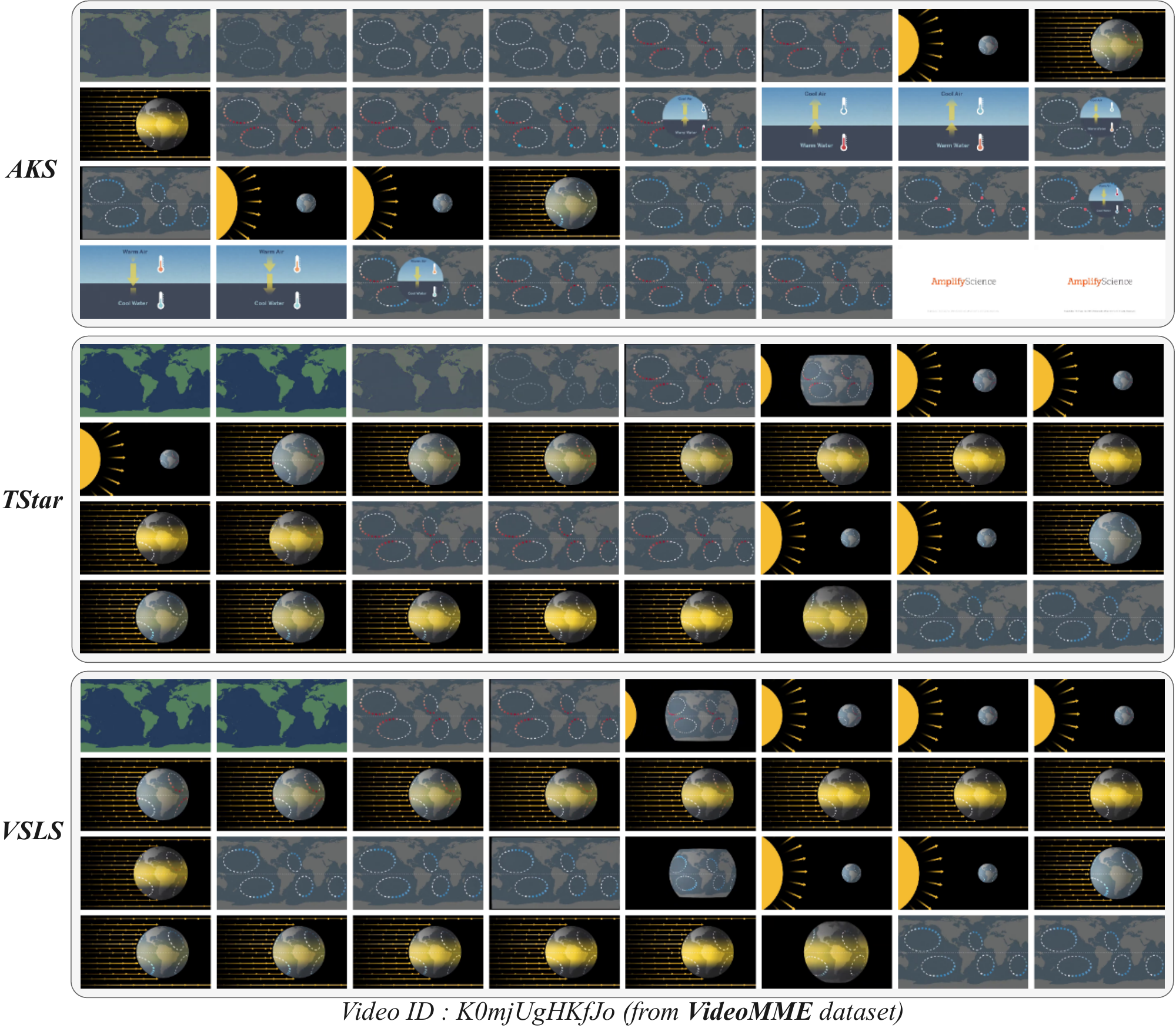} 
    \caption{\textbf{Case Study 2: `Visual Echoes' in a Scientific Animation Video.} This example (Video ID: K0mjUgHKfJo) shows that even in educational content with distinct conceptual phases, all three selectors produce highly repetitive frames of the Earth's rotation and its position relative to the sun.}
    \label{fig:appendix_echoes_earth}
\end{figure}

\begin{figure}[h!]
    \centering
    \includegraphics[width=\columnwidth]{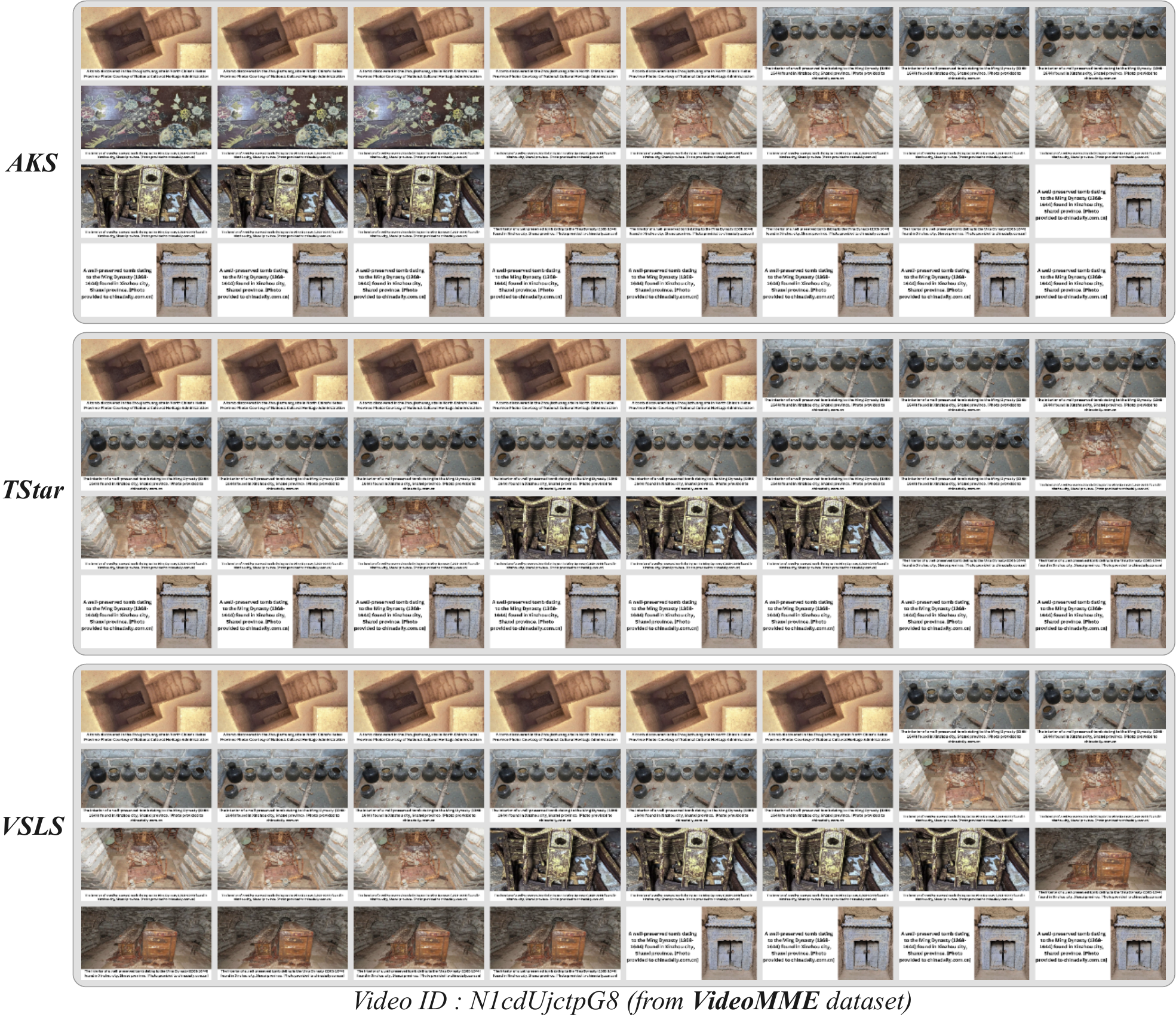} 
    \caption{\textbf{Case Study 3: `Visual Echoes' in a Real-World Video.} This example (Video ID: N1cdUjctpG8) demonstrates the prevalence of redundancy in complex, real-world footage. Despite challenging lighting and rich details, all selectors still select numerous near-identical shots of the same artifacts.}
    \label{fig:appendix_echoes_tomb}
\end{figure}

To further demonstrate that `visual echoes' are a pervasive challenge not limited to a specific video type, this section provides additional case studies. While our main paper (Figure 2) uses a simple animation to clearly illustrate the issue, here we show that the same pattern of severe frame redundancy occurs in more complex scenarios, including a scientific animation (Figure~\ref{fig:appendix_echoes_earth}) and a real-world video with significant visual detail (Figure~\ref{fig:appendix_echoes_tomb}).

%
%

\section{Additional Details on Experimental Setup}
\label{sec:appendix_exp_details}

\subsection{Dataset Subsets for Fair Evaluation}
\label{sec:appendix_dataset_curation}

To ensure maximum transparency, reproducibility, and fairness, our experiments were conducted on consistent, curated subsets of the official \lvb and \mme benchmarks. Our initial setup involved baselines like \basevsstar, whose original implementation filtered out videos that were unreadable or caused processing errors. This resulted in a clean, stable subset of 646 samples for \lvb and 2657 samples for \mme. To maintain a strict apples-to-apples comparison, we adopted these exact same subsets for \textbf{all} methods reported in our paper, including our main baseline \baseaksstar and our proposed \methodgraphspace framework. This approach ensures that all performance differences are attributable solely to the methods themselves, not variations in data quality.

%
%
\section{Note on the \baseaksstar Baseline Adaptation}
\label{sec:appendix_baseline_modification}

In our main experiments, we use \baseaksstar as our primary sophisticated baseline. It is important to clarify the distinction between the original \baseaks~\cite{tang2025adaptive} framework and our adapted \baseaksstar version, and to justify this choice.

\paragraph{Rationale for Adaptation.}
Unlike other selectors such as \basetstar and \basevs that employ a direct Top-K selection, the original \baseaks utilizes a more sophisticated \textbf{dynamic sampling algorithm}. This algorithm recursively allocates frame budgets to temporal segments based on score distributions. While innovative, this dynamic approach does not guarantee a fixed number of output frames (e.g., it may return 6 frames when a budget of 8 is requested), making rigorous, fixed-budget comparisons challenging. To ensure a fair and reproducible evaluation protocol, we therefore modify its final stage to a standard Top-K selection. Crucially, this choice is empirically supported by the ablation studies within the original \baseaks paper itself. Their experiments show that a standard Top-K selection (denoted `TOP') is the best performing alternative, achieving results highly competitive with their full adaptive method (`ADA') (e.g., 62.4\% vs. 62.7\% on \lvb). This confirms that our \baseaksstar adaptation, based on Top-K selection, serves as a powerful and faithful representation of the original method's capabilities for the purpose of a strong baseline comparison.

\paragraph{Validity of the `Visual Echoes' Motivation.}
This modification does not weaken our core claim regarding the prevalence of `visual echoes'. The problem of frame redundancy is not an artifact of simple Top-K selection. As visually demonstrated in our qualitative examples (Figure~\ref{fig:appendix_echoes_earth}, Figure~\ref{fig:appendix_echoes_tomb} and Section~\ref{sec:appendix_additional_echoes}), `visual echoes' are a fundamental challenge rooted in the temporal nature of video. The original, more complex dynamic sampling of \baseaks also suffers from this issue because if a temporal segment with high visual similarity receives a high score, the algorithm will still allocate budget to select multiple, similar frames from within that segment. This reinforces our central argument: regardless of the final selection strategy, the `visual echoes' problem persists, universally motivating the need for a dedicated refinement layer like our \methodshort framework.

%
%
\section{Further Implementation Details}
\label{sec:appendix_implement_deatils} 

This section provides in-depth details about our implementation to ensure full reproducibility of our work.

\subsection{\methodshort Algorithm Details}
Our \methodfullname algorithm is implemented in Python utilizing the scikit-learn~\cite{scikit-learn} library. The following implementation specifics directly complement the Methodology section of the main paper.

\noindent\textbf{Clustering Parameters.}
For the core clustering step, we employ the \texttt{AgglomerativeClustering} class from scikit-learn. The \texttt{linkage} parameter is set to `\texttt{average}', which means the distance between two clusters is defined as the average of the distances between all pairs of samples, with one sample from each cluster.

\noindent\textbf{Small Cluster Refinement.}
To enhance the robustness of our clustering, we implement a refinement step (\texttt{refine\_clusters} in our script). After the initial clustering, any cluster containing fewer than two frames is considered unstable. Such singleton clusters are merged into their nearest neighboring cluster, where proximity is determined by the average visual cosine distance between the frames of the two clusters. This prevents trivial clusters and ensures a more meaningful grouping.

\noindent\textbf{Adaptive Threshold Calculation.}
As mentioned in the main paper, the \texttt{distance\_threshold} for clustering is determined adaptively. Specifically, after calculating all pairwise visual cosine distances, we fit a Gaussian Kernel Density Estimator (KDE) to their distribution. The threshold is then calculated as $\tau = p + 0.15$, where $p$ is the distance value corresponding to the peak of the density function. The constant offset of 0.15 is an empirically chosen value designed to prevent the clustering from being overly conservative (i.e., creating too many small clusters) and to encourage the merging of closely related `visual echoes'.

\subsection{Prompt Structure and MLLM Inference}

The structure of the prompts sent to the MLLM is critical for achieving consistent and reproducible results. In this section, we detail the two primary prompt templates used in our framework: one for our versatile semantic graph generation (Case 2), and one for the final downstream Video-QA task.

\subsection{Prompt Template for Semantic Graph}

\begin{figure}[h!]
\centering
\includegraphics[width=\columnwidth]{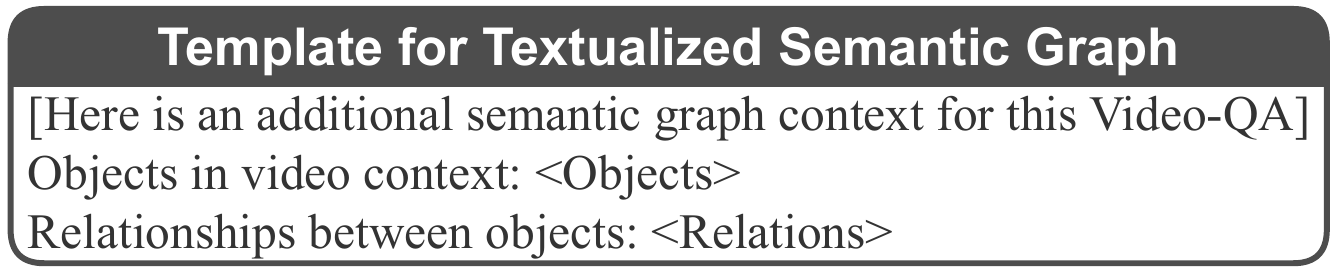}
\caption{\textbf{The structural template for the textualized semantic graph.} This block is programmatically generated based on the query and inserted into the final downstream prompt. It serves to provide the MLLM with high-level, pre-digested semantic context to guide its reasoning over the visual frames.}
\label{fig:appendix_graph_template}
\end{figure}

\noindent Our framework generates a concise, structured block of text, termed the \textbf{Textual Semantic Graph}, which is programmatically injected into the final prompt for the downstream MLLM. This graph provides high-level semantic context, summarizing the key entities and their inter-relationships as inferred from the query. Figure~\ref{fig:appendix_graph_template} illustrates the structural template of this generated text block.

\begin{figure*}[h!]
\centering
\includegraphics[width=\textwidth]{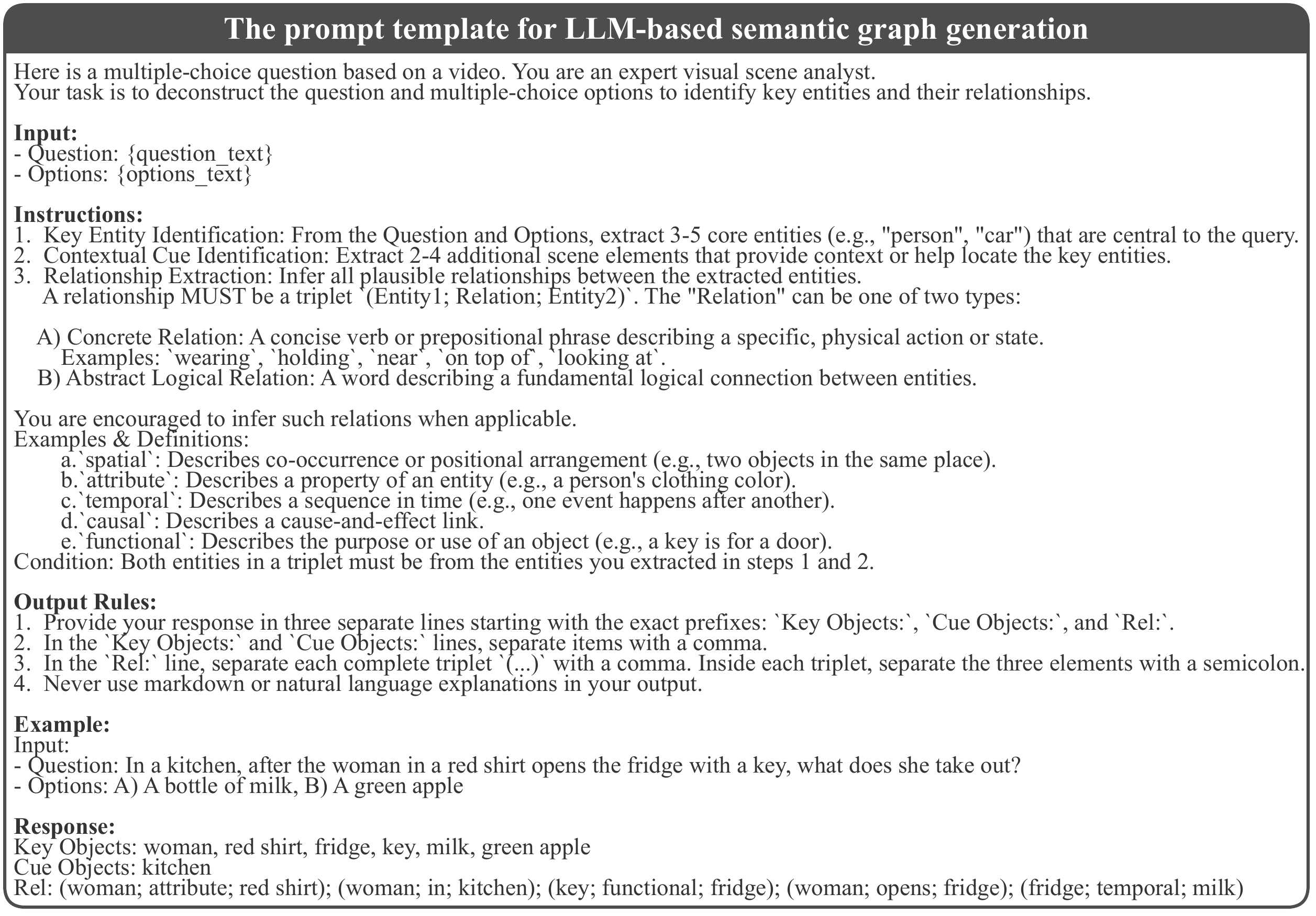}
\caption{\textbf{The full prompt template for LLM-based semantic graph generation.} This prompt leverages persona, chain-of-thought, and open-ended relation extraction to guide the LLM in deconstructing the query.}
\label{fig:prompt_semantic_graph}
\end{figure*}

\noindent\textbf{Prompt for Semantic Graph Generation.}
For our low-cost, universally applicable semantic graph generation strategy (Case 2), we employ a sophisticated prompt designed to elicit rich, structured information from the MLLM based solely on textual input. As detailed in Figure~\ref{fig:prompt_semantic_graph}, the prompt is structured using several advanced prompting techniques:
\begin{itemize}
    \item \textbf{Persona Prompting}: We assign the MLLM the role of an ``expert visual scene analyst'' to activate its advanced reasoning capabilities.
    \item \textbf{Chain-of-Thought Instructions}: The prompt provides a clear, step-by-step thinking process (Key Entity Identification, Contextual Cue Identification, Relationship Extraction) for the MLLM to follow.
    \item \textbf{Open-ended yet Structured Relation Extraction}: Crucially, our prompt encourages flexibility. It explicitly defines two categories of relations—Concrete (e.g., verbs, prepositions) and Abstract Logical (e.g., `spatial', `causal')—and provides clear examples and definitions for each. This "dual-track" design allows for the capture of a much richer and more nuanced set of semantic connections while maintaining a structured output.
    \item \textbf{Few-Shot Example}: A comprehensive example is provided to demonstrate the desired input-output format and the expected quality of the extracted graph.
\end{itemize}
The full prompt is shown in Figure~\ref{fig:prompt_semantic_graph}.

\begin{figure}[h!]
\centering
\includegraphics[width=\columnwidth]{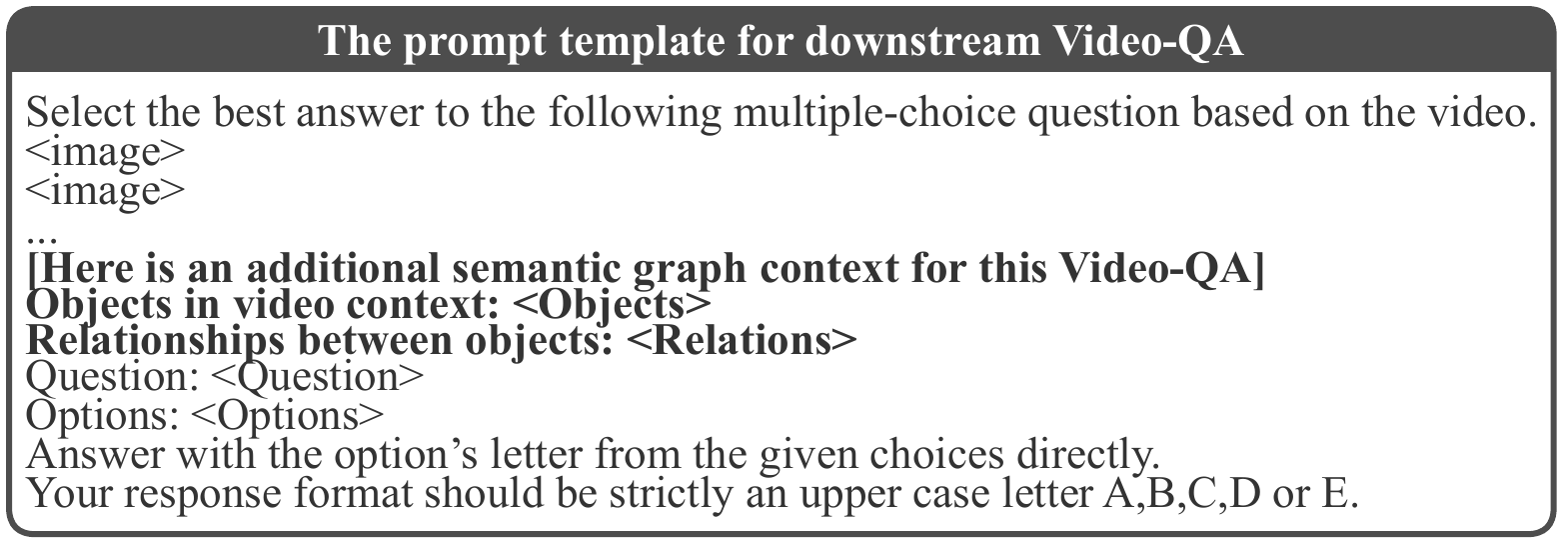}
\caption{\textbf{The full MLLM prompt template used for the final Video-QA task.} It integrates image placeholders, the optional semantic graph, and the QA content into a single, direct query.}
\label{fig:prompt_qa}
\end{figure}

\noindent\textbf{Prompt for Downstream Video-QA.}
For the final question-answering stage, we utilize a clean, direct, and robust prompt template, detailed in Figure~\ref{fig:prompt_qa}. This prompt is designed for maximum compatibility across different MLLMs. It clearly presents the visual evidence (via `\textless image\textgreater' placeholders), followed by the optional textual semantic graph, and finally the question and multiple-choice options. The instructions are direct and unambiguous, strictly constraining the model to output a single uppercase letter corresponding to its chosen answer. This minimalist design minimizes prompt engineering sensitivity and ensures that performance differences are attributable to the quality of the visual and semantic inputs, rather than variations in prompt interpretation.

%
%

\section{Hyperparameter and Component Analysis}
\label{sec:appendix_ablation_deatils} 

\subsection{Ablation Study on Prompt Formulation}
The formulation of the prompt, particularly how the semantic graph is textualized and how instructions are given to the MLLM, can significantly impact performance. To determine the optimal prompt structure for our main experiments, we conducted an ablation study on several variants. All experiments were performed on the LongVideoBench dataset with the GPT-4o model, starting from Top-8 keyframes.

\subsubsection{Prompt Component Variants}
We tested combinations of two \texttt{graph\_context} formats and two \texttt{system\_prompt} formats:

\noindent\textbf{Graph Context Formats.}
\begin{itemize}
    \item \textbf{Concise Triplet (G1):} Our final choice. This format is simple and structured, directly presenting nodes and raw triplet relationships (e.g., \texttt{(object1, relation, object2)}).
    \item \textbf{Verbose Natural Language (G2/G3):} These formats attempt to convert the graph into more human-like sentences (e.g., ``object1 appears with object2'').
\end{itemize}

\noindent\textbf{System Prompt Formats.}
\begin{itemize}
    \item \textbf{Direct Prompt (P1):} Our final choice. This is a concise, direct instruction for the QA task.
    \item \textbf{Instructional Prompt (P2):} This is a more verbose prompt that assigns an ``expert'' persona to the MLLM and provides detailed step-by-step guidelines.
\end{itemize}

\subsubsection{Results and Analysis}
The results of our prompt ablation are presented in Table~\ref{tab:prompt_ablation}.

\noindent\textbf{A ``Less is More'' Phenomenon in Prompting.}
Interestingly, this ablation study reveals a microcosm of our paper's central `Less is More' theme, but applied to the prompt engineering space. As shown in Table~\ref{tab:prompt_ablation}, while more verbose and complex combinations like (G1, P2) or (G3, P2) can achieve marginally higher peak accuracy on specific video lengths (e.g., Short videos), this gain is inconsistent and comes at the cost of stability and increased token counts. The simpler, more direct (G1, P1) format demonstrates a superior balance across all metrics. This suggests that MLLMs can suffer from a form of \textbf{``instructional noise''} or ``prompt dilution,'' where overly elaborate instructions can sometimes obscure the core task, mirroring how excessive visual frames can cause ``context dilution.'' Based on this finding, our analysis led us to select the combination of \textbf{Concise Triplet graph (G1)} and \textbf{Direct Prompt (P1)} for all main experiments, guided by the following key reasons:

\begin{itemize}
    \item \textbf{Robust and Stable Performance:} While other combinations, such as (G1, P2), achieve marginally higher accuracy on certain video lengths, the (G1, P1) combination demonstrated the most stable and consistent high performance across all our preliminary and main experiments. As noted in our experimental logs, other verbose combinations frequently yielded unexpectedly low results in some runs, indicating a lack of robustness. The (G1, P1) setting, in contrast, reliably produced strong results.
    
    \item \textbf{Token Efficiency:} The G1 and P1 formats are significantly more concise than their verbose counterparts. This results in a lower token count for each query, which directly aligns with our paper's core objective of maximizing token efficiency.
    
    \item \textbf{Simplicity and Generalizability:} The direct, structured format of (G1, P1) provides clear instructions to the MLLM without excessive ``prompt engineering.'' We believe this simpler format is more likely to generalize well across different MLLMs, as it relies on fundamental instruction-following capabilities rather than sensitivity to nuanced persona-based instructions.
\end{itemize}

In conclusion, our chosen (G1, P1) prompt formulation represents the best trade-off between performance, stability, and token efficiency, making it the most suitable choice for our proposed method.

\begin{table}[h!]
\centering
\caption{\textbf{Ablation study on prompt formulation. Results are averaged over multiple runs.} Our chosen combination (G1, P1) is highlighted in \textbf{bold}.}
\begin{tabularx}{\columnwidth}{l l C C C} 
\toprule
\textbf{Graph} & \textbf{Prompt} & \multicolumn{3}{c}{\textbf{Avg. Accuracy (\%)}} \\
\cmidrule(lr){3-5}
\textbf{Format} & \textbf{Format} & \textbf{Long} & \textbf{Medium} & \textbf{Short} \\
\midrule
\textbf{G1 (Ours)} & \textbf{P1 (Ours)} & \textbf{47.0} & 52.5 & 69.0 \\
G2 & P1 & 46.1 & 51.7 & 66.4 \\
G1 & P2 & 46.7 & \textbf{52.9} & \textbf{70.7} \\
G2 & P2 & 45.9 & 49.9 & 69.0 \\
G3 & P2 & 44.9 & 51.2 & \textbf{70.7} \\
\bottomrule
\end{tabularx}
\label{tab:prompt_ablation}
\end{table}

\subsection{Analysis of Hyperparameters}
\label{sec:appendix_hyper_analysis} 

\begin{figure}[h!]
\centering
\includegraphics[width=\columnwidth]{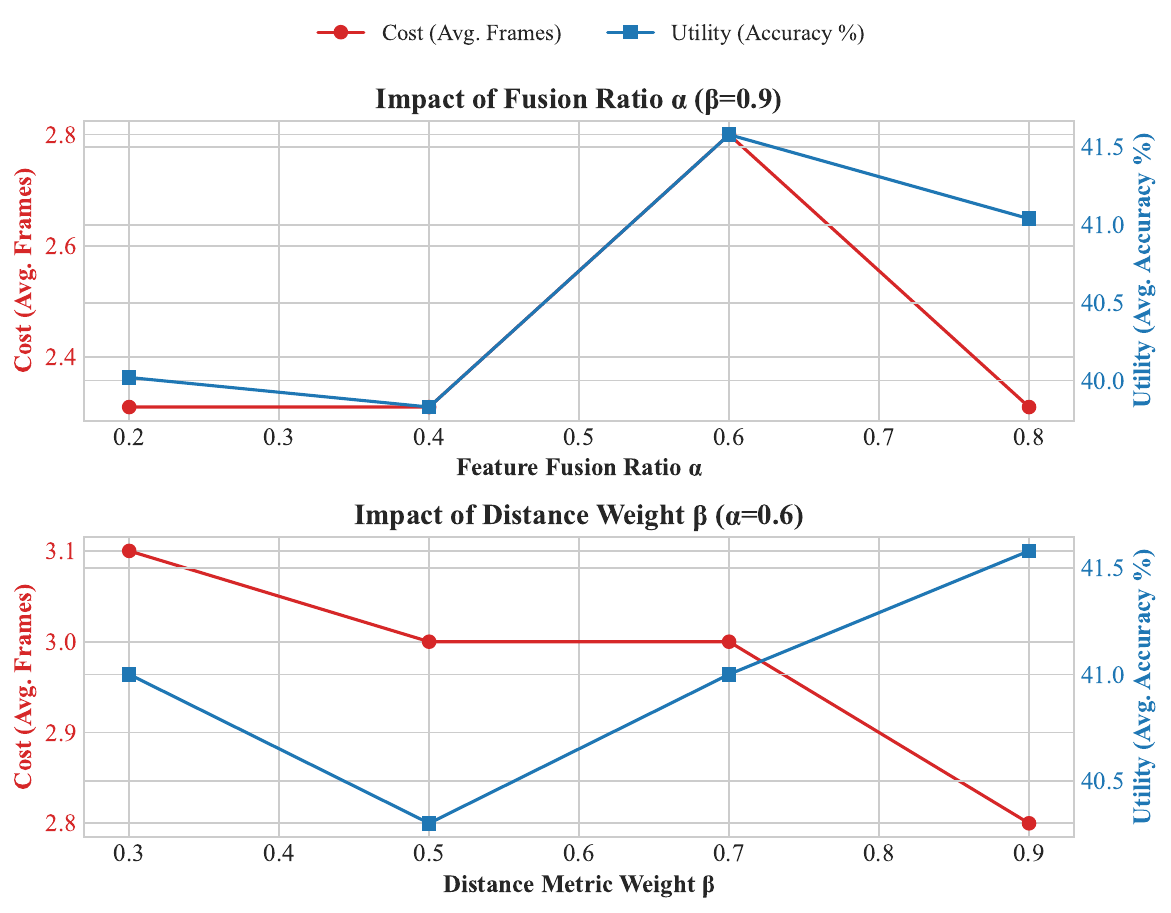} 
\caption{\textbf{Cost-Utility trade-off analysis for hyperparameters $\alpha$ and $\beta$.} \textbf{Cost} (left y-axis, red) is measured by the average number of output frames, while \textbf{Utility} (right y-axis, blue) is measured by the average QA accuracy on a subset of \lvb. The plots demonstrate that our parameters are tunable knobs: users can adjust them to prioritize either higher accuracy (e.g., lower $\beta$) or maximum efficiency (e.g., higher $\beta$). Our default choice (`$\alpha=0.6, \beta=0.9$') represents a strong balance.}
\label{fig:appendix_hyper_tradeoff} 
\end{figure}

This section provides the detailed data and analysis supporting our hyperparameter selection, ensuring the reproducibility of our work. Our method's core hyperparameters, the feature fusion ratio $\alpha$ and the distance metric weight $\beta$, were determined through a systematic sensitivity analysis. This analysis reveals a clear cost-utility trade-off, allowing our framework to be tuned to prioritize either performance (utility) or efficiency (cost). All experiments were performed on the \lvb dataset using the Qwen2.5-VL-7B-Instruct model, starting from \basevsstar Top-8 keyframes. The results, averaged over multiple runs, are visualized in Figure~\ref{fig:appendix_hyper_tradeoff} and detailed with exact values in Table~\ref{tab:hyperparameter_analysis_supp}.

\begin{table}[h!]
\centering
\caption{\textbf{Detailed sensitivity analysis of hyperparameters $\alpha$ and $\beta$.} Results are averaged over multiple runs on the \lvb dataset using the Qwen2.5-VL-7B-Instruct model (from Top-8). Our chosen parameters ('$\alpha=0.6, \beta=0.9$') are highlighted in \textbf{bold}.}
\begin{tabularx}{\columnwidth}{C C C}
\toprule
\textbf{Hyperparameter Setting} & \textbf{Avg. Accuracy (\%)} & \textbf{Avg. Frames} \\
\bottomrule
\multicolumn{3}{c}{\textit{Analysis of Feature Fusion Ratio $\alpha$ (fixing $\beta=0.9$)}} \\
\hline
$\alpha$ = 0.2 & 40.02 & \textbf{2.31} \\
$\alpha$ = 0.4 & 39.83 & \textbf{2.31} \\
\textbf{$\alpha$ = 0.6 (Ours)} & \textbf{41.58} & 2.80 \\
$\alpha$ = 0.8 & 41.04 & \textbf{2.31} \\
\bottomrule
\multicolumn{3}{c}{\textit{Analysis of Distance Metric Weight $\beta$ (fixing $\alpha=0.6$)}} \\
\hline
$\beta$ = 0.3 & 41.00 & 3.10 \\
$\beta$ = 0.5 & 40.30 & 3.00 \\
$\beta$ = 0.7 & 41.00 & 3.00 \\
\textbf{$\beta$ = 0.9 (Ours)} & \textbf{41.58} & \textbf{2.80} \\
\bottomrule
\end{tabularx}
\label{tab:hyperparameter_analysis_supp}
\end{table}

\noindent\textbf{Analysis and Selection Strategy.}
As visualized in Figure~\ref{fig:appendix_hyper_tradeoff} and detailed in Table~\ref{tab:hyperparameter_analysis_supp}, our analysis reveals a clear cost-utility trade-off. For the feature fusion ratio $\alpha$ (top plot), the results show that $\alpha=0.6$ achieves the highest average accuracy of 41.58\%, suggesting an optimal balance between visual and semantic features, albeit at a slightly higher frame cost. For the distance metric weight $\beta$ (bottom plot), a higher value (e.g., 0.9) leads to more aggressive clustering (fewer frames, as shown by the red line) and, in this case, also yields the highest accuracy. This demonstrates that the hyperparameters of \methodshort are not arbitrary fixed values, but rather tunable ``knobs'' that control the cost-utility balance. Based on this systematic analysis, we selected the configuration of \textbf{$\alpha=0.6$} and \textbf{$\beta=0.9$} for all main experiments, as it consistently provided the best overall performance and a favorable efficiency-accuracy balance in our development set.

\subsection{Representative Frame Selection Strategy}

A crucial step in our \methodshort algorithm is selecting a single representative frame from each generated cluster. To determine the most effective and robust approach, we conducted a comprehensive ablation study comparing three distinct strategies. The experiments were performed on the \lvb dataset with the GPT-4o model, starting from Top-8 keyframes provided by \basevsstar, and the results reported are averaged over multiple runs to ensure reliability. This analysis focuses on the `\methodshort only' setting (without the semantic graph) to purely evaluate the quality of the selected visual information.

\begin{itemize}
    \item \textbf{Score-based (Highest \basevsstar Score):} This strategy leverages external information by selecting the frame from each cluster that has the highest initial relevance score assigned by the upstream \basevsstar selector.
    \item \textbf{Centroid-based (Visual Centroid):} This strategy is self-contained. It selects the ``centroid frame'', the frame with the minimum average feature distance to all other frames within the same cluster, making it the most visually representative.
    \item \textbf{Relevance-based (Highest Query Similarity):} This strategy introduces task-specific guidance by selecting the frame with the highest CLIP similarity score to the `Question + Options' text prompt.
\end{itemize}

\begin{table*}[t!]
\centering
\caption{\textbf{Ablation study on representative frame selection strategies.} The experiment was conducted in the `\methodshort only' setting to purely evaluate the quality of the selected visual information. Results are averaged over multiple runs. The Centroid-based strategy demonstrates the best performance.}
\begin{tabularx}{\textwidth}{lll >{\centering\arraybackslash}X cc}
\toprule
\multicolumn{4}{c}{\textbf{Experimental Settings}} & \multicolumn{2}{c}{\textbf{Averaged Results}} \\
\cmidrule(lr){1-4} \cmidrule(lr){5-6}
\textbf{Dataset} & \textbf{Model} & \textbf{Frame Source} & \textbf{Selection Strategy} & \textbf{Avg. Acc (\%)} & \textbf{Avg. Frames} \\
\midrule
\multirow{3}{*}{\lvb} & \multirow{3}{*}{GPT-4o} & \multirow{3}{*}{\basevsstar (Top-8)} & Score-based & 47.14 & 2.1 \\
& & & \textbf{Centroid-based} & \textbf{48.53} & \textbf{2.1} \\
& & & Relevance-based & 46.67 & 2.2 \\
\bottomrule
\end{tabularx}
\label{tab:selection_strategy}
\end{table*}

As shown in Table~\ref{tab:selection_strategy}, the averaged results over multiple runs confirm that the \textbf{Centroid-based strategy} holds a clear advantage. It achieves the highest overall accuracy (48.53\%), outperforming both the Score-based and Relevance-based approaches. The superiority of the Centroid strategy is theoretically sound, especially in the absence of a semantic graph. By selecting the frame closest to the cluster's feature-space center, it guarantees the most visually representative frame is chosen, which is the most robust way to minimize information loss during the pruning process. This self-contained logic is arguably more generalizable than relying on external scores (from \basevsstar or CLIP), which might be noisy or biased. Based on these comprehensive findings, we adopted the \textbf{Centroid-based strategy} for all main experiments reported in this paper.

%
%

\section{Robustness on Event Dense Short Videos}
\label{sec:appendix_holmes}
Our framework primarily targets long videos where sparse sampling inevitably yields severe visual redundancy. To rigorously evaluate its robustness in out of distribution scenarios, we conducted a stress test on the Video-HOLMES~\cite{cheng2025video} benchmark. This dataset consists of short, event dense videos (typically 1 to 5 minutes) where a 32 frame extraction represents dense sampling with minimal inherent redundancy. We evaluated the Qwen2.5-VL-7B-Instruct model on a stratified subset of 900 videos. The results in Table~\ref{tab:holmes} demonstrate that our method remains highly robust. Even with minimal redundancy to prune, the \methodshort module successfully filters out subtle noise, improving accuracy by 2.03\% while reducing the frame count to 4.1. The integration of the semantic graph further elevates the accuracy to 30.14\%, achieving a total absolute improvement of 3.47\% over the \baseaksstar baseline. This confirms that our framework safely compresses visual information and provides a reliable semantic reasoning scaffold even in highly dynamic contexts.

\begin{table}[h!]
    \centering
    \footnotesize
    \setlength{\tabcolsep}{6pt}
    \caption{\textbf{Stress test on the Video-HOLMES benchmark.} Our method maintains robustness and improves accuracy on event dense short videos.}
    \begin{tabular}{l l c c c}
        \toprule
        \textbf{Model} & \textbf{Method} & \textbf{Avg. Fr.} & \textbf{Acc (\%)} & \textbf{vs. Base} \\
        \midrule
        \multirow{3}{*}{Qwen2.5-VL} & \baseaksstar (Base) & 32.0 & 26.67 & / \\
         & \methodshort only & 4.1 & 28.70 & \textcolor{green}{+2.03} \\
         & \methodgraphspace & \textbf{4.1} & \textbf{30.14} & \textcolor{green}{\textbf{+3.47}} \\
        \bottomrule
    \end{tabular}
    \label{tab:holmes}
\end{table}

\section{Reliability of the Semantic Graph}
\label{sec:appendix_hallucination}
A potential concern regarding the use of an LLM generated semantic graph is the introduction of textual hallucinations. To quantitatively assess the reliability of our generated context, we conducted a rigorous manual audit of 100 randomly sampled graphs from the VIDEOMME dataset. The audit revealed an exceptionally high degree of fidelity, achieving an entity extraction accuracy of 98\% and a relationship extraction accuracy of 96\%. Qualitative analysis showed that the minimal inaccuracies observed (approximately 4\%) were predominantly relevance errors (e.g., extracting a peripheral background object) rather than factual fabrications. This high reliability is deliberately engineered into our framework: by constraining the LLM to perform a structural decomposition of the textual query rather than open ended generation, we strictly limit the generative space. Furthermore, the downstream MLLM utilizes the sampled visual frames as primary evidence, inherently filtering out minor semantic noise.

%
%
\begin{figure}[t!]
    \centering
    \includegraphics[width=\columnwidth]{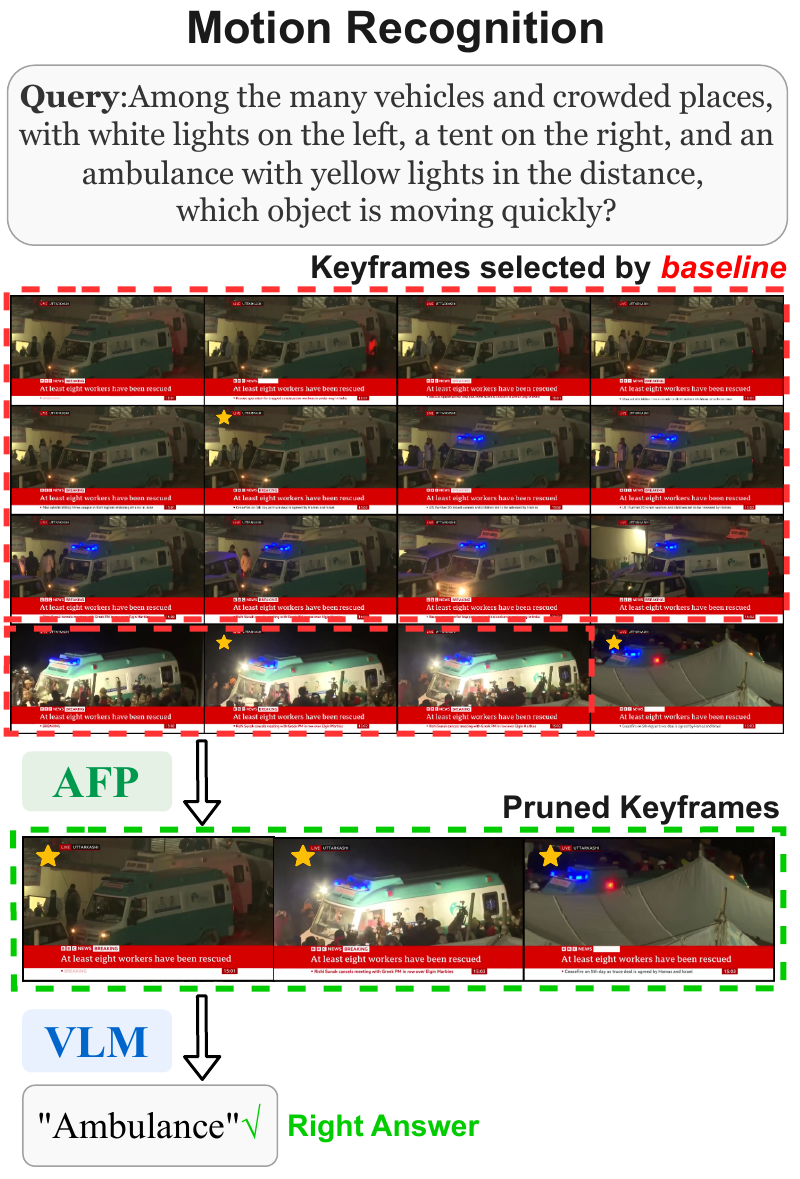} 
    \caption{Qualitative example for a \textbf{motion recognition task} (Video ID: Z00vWImw1KQ). The query asks to identify a moving object. The baseline \basevsstar{} selects 16 frames with high redundancy. Our \methodshort{} algorithm prunes this set to just 3 frames that preserve the key motion cues, enabling the MLLM to derive the correct answer.}
    \label{fig:qualitative_motion} 
\end{figure}



\section{Qualitative Analysis and Limitations}
\label{sec:qualitative_analysis}
\subsection{Additional Qualitative Examples}

To demonstrate the versatility and robustness of our \methodfullname{} method, this section provides qualitative examples from diverse Video-QA tasks. These cases illustrate how \methodshort{} effectively handles various forms of visual redundancy while preserving the crucial information needed to answer different types of queries.
Figure~\ref{fig:qualitative_motion} presents a challenging task requiring the identification of a fast-moving object based on a descriptive, distractor-rich query. The baseline \basevsstar{} method, aiming for comprehensive coverage, selects 16 keyframes. However, these frames are plagued by severe `visual echoes': the first twelve frames are nearly identical, capturing a static view of an ambulance. This flood of repetitive information risks overwhelming the MLLM. Our \methodshort{} algorithm excels in such scenarios. It correctly identifies and merges the redundant static and slightly shifted shots into distinct representative frames. Consequently, the input is drastically reduced to a highly efficient set of just 3 frames. By capturing the ambulance's initial state, a subtle movement, and the surrounding context, these frames preserve the critical temporal cues needed to infer motion. This compact visual input, augmented by the textual semantic graph, provides a sufficient, non-redundant context for the MLLM to correctly identify the ``Ambulance''.

\subsection{Qualitative Analysis of Limitations}
\label{sec:appendix_limitations}

To provide a balanced perspective on our framework, this section analyzes its inherent limitations through two representative failure cases, which also highlight avenues for future research.

\begin{figure}[h!]
\centering
\includegraphics[width=\columnwidth]{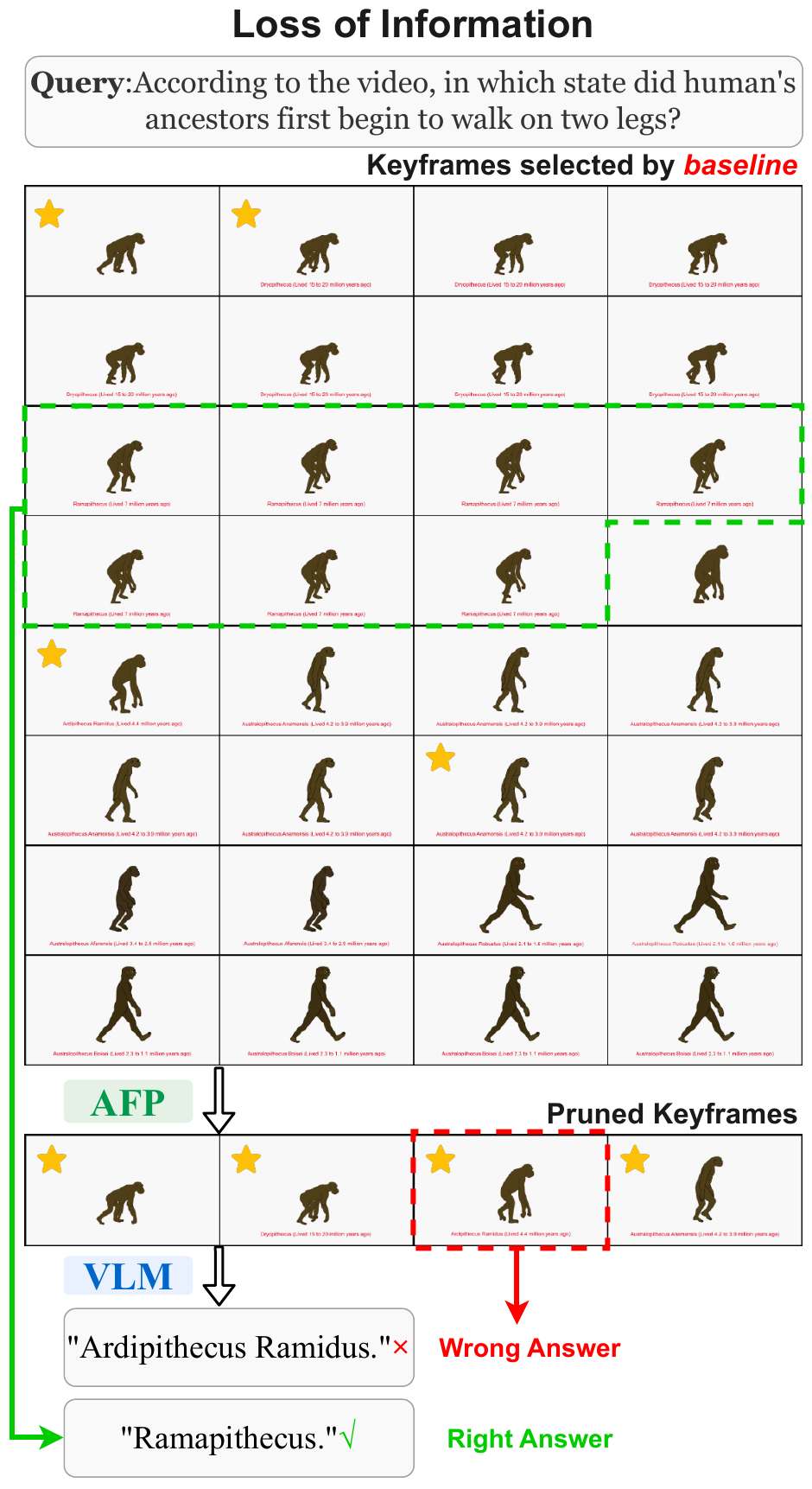}
\caption{An analysis of a failure case caused by \textbf{information loss during pruning}. The initial keyframes from \basevsstar contain the evidence for the correct answer ``Ramapithecus'' (highlighted in green). However, due to high visual similarity with other frames, our \methodshort algorithm incorrectly prunes this crucial frame, leading the VLLM to a wrong answer based on the remaining evidence.}
\label{fig:failure_case1}
\end{figure}

\noindent\textbf{Loss of Fine-Grained Information during Pruning.}
The first failure case (Figure~\ref{fig:failure_case1}) highlights a trade-off inherent to our global feature-based pruning. While correctly identifying the high visual similarity between frames depicting ``Ramapithecus'' and ``Ardipithecus Ramidus'', our \methodshort algorithm merges them into a single cluster and inadvertently discards the frame containing the correct evidence. This reveals a primary limitation: a reliance on global features can be insensitive to subtle but semantically decisive local details, such as small pieces of text. This points towards a clear avenue for future improvement by integrating more localized feature extractors, such as Optical Character Recognition (OCR) models, into the clustering process.

\begin{figure}[h!]
\centering
\includegraphics[width=\columnwidth]{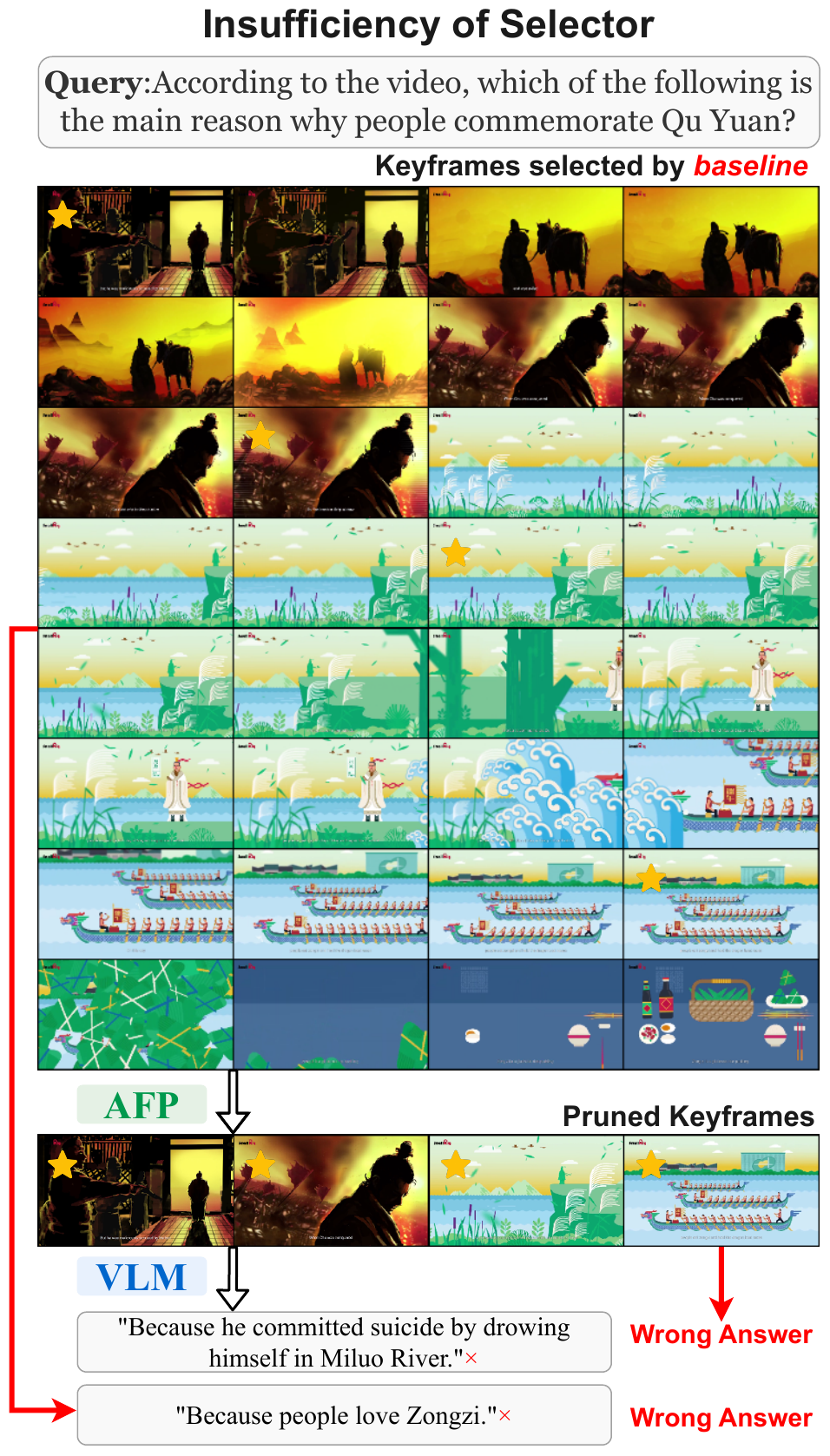}
\caption{An analysis of a failure case caused by the \textbf{insufficiency of the upstream selector}. The question requires a global understanding of a narrative. The initial 32 frames from \basevsstar are already insufficient to answer correctly. Our \methodshort method, while efficiently pruning the input, cannot recover the missing information, leading to an inevitable failure by the VLLM. This illustrates the performance ceiling imposed by the initial keyframe selection.}
\label{fig:failure_case2}
\end{figure}

\noindent\textbf{Ceiling Effect from Upstream Selector.}
The second case (Figure~\ref{fig:failure_case2}) perfectly exemplifies our framework's role as a post-processing refinement layer. Here, the query requires a deep narrative understanding that is absent even in the initial 32 frames provided by the upstream selector. Our \methodshort module, while correctly pruning the insufficient input, cannot recover this missing information, leading to an unavoidable failure. This case demonstrates that our framework's performance is fundamentally capped by the quality of the initial keyframe set, as its role is to prune, not to discover. This limitation highlights a broader challenge for all sparse sampling methods when dealing with questions that demand global, narrative-level reasoning.

\section{Future Directions}
\label{sec:appendix_future_directions}

The preceding failure case analysis illuminates two promising avenues for future research. First, to address the loss of fine-grained details, our framework's reliance on global visual features can be overcome by developing an \textbf{adaptive feature fusion mechanism}. This system would dynamically switch to more \textbf{localized, patch-based representations} or integrate Optical Character Recognition (OCR) models to re-evaluate clusters where subtle details are semantically decisive. Ultimately, both limitations point towards a unified solution: the development of a \textbf{`Multimodal Semantic Graph'}. By encapsulating not only textual relationships but also key visual motifs, audio cues, and explicit textual evidence from OCR, this compact representation would shift the paradigm from mere frame selection towards a more holistic video comprehension. This would provide a more robust and token-efficient foundation for reasoning, paving the way for more scalable video AI.


\end{document}